\documentclass[preprint,12pt]{elsarticle}

\usepackage[margin=0.7in]{geometry}
\usepackage{framed,multirow}
\usepackage{booktabs}
\usepackage{tabularray}
\UseTblrLibrary{booktabs}
\usepackage{boldline} %$Bold lines for tables
\usepackage{fontawesome5}
\usepackage{amssymb,amsmath}
\usepackage{romannum}
\AtBeginDocument{\pagenumbering{arabic}}
\usepackage{pifont}

\usepackage{latexsym}
\usepackage{mathrsfs}
\usepackage[ruled,vlined]{algorithm2e}
\usepackage{soul}
\usepackage{url}
\usepackage[dvipsnames,table,xcdraw, svgnames,x11names]{xcolor}
\usepackage{makecell}

\usepackage{blindtext}
\usepackage{graphicx}

\usepackage[listings,skins,breakable]{tcolorbox}
\usepackage[colorlinks]{hyperref}
\usepackage[printonlyused,withpage]{acronym}
\usepackage[nameinlink]{cleveref}
\usepackage{caption}
\usepackage{subcaption}

\newcommand{\crossmark}{\textcolor{red}{\ding{55}}} % Red cross mark
\newcommand{\checkmarks}{\textcolor[RGB]{0,153,76}{\checkmark}} % Green check mark
\definecolor{newcolor}{rgb}{.8,.349,.1}
\definecolor{maroon}{cmyk}{0,0.87,0.68,0.32}
\definecolor{lightorange}{rgb}{1,0.753,0.478}

\graphicspath{{./}}
\begin{document}

\begin{frontmatter}

\title{Medical Image Classification with KAN-Integrated Transformers and Dilated Neighborhood Attention}%

\author[1]{Omid Nejati Manzari} \corref{cor1}
\author[1]{Hojat Asgariandehkordi}
\author[1]{Taha Koleilat}
\author[2]{Yiming Xiao}
\author[1]{Hassan Rivaz}

\cortext[cor1]{Corresponding author: omid.nejatimanzari@mail.concordia.ca}

%\address[1]{Independent Researcher, Tehran, Iran}
\address[1]{Department of Electrical and Computer Engineering, Concordia University, Montreal, Canada}
\address[2]{Department of Computer Science and Software Engineering, Concordia University, Montreal, Canada}

% \received{1 May 2013}
% \finalform{10 May 2013}
% \accepted{13 May 2013}
% \availableonline{15 May 2013}
% \communicated{S. Sarkar}

\begin{abstract} Convolutional networks, Transformers, hybrid models, and Mamba-based architectures have demonstrated strong performance across various medical image classification tasks. However, these methods were primarily designed to classify clean images using labeled data. In contrast,  real-world clinical data often involve image corruptions that are unique to multi-center studies and stem from variations in imaging equipment across manufacturers.
In this paper, we introduce the Medical Vision Transformer (MedViTV2), a novel architecture incorporating Kolmogorov-Arnold Network (KAN) layers into the Transformer architecture for the first time, aiming for generalized medical image classification. We have developed an efficient KAN block to reduce computational load while enhancing the accuracy of the original MedViT. Additionally, to counteract the fragility of our MedViT when scaled up, we propose an enhanced Dilated Neighborhood Attention (DiNA), an adaptation of the efficient fused dot-product attention kernel capable of capturing global context and expanding receptive fields to scale the model effectively and addressing feature collapse issues. Moreover, a hierarchical hybrid strategy is introduced to stack our Local Feature Perception and Global Feature Perception blocks in an efficient manner, which balances local and global feature perceptions to boost performance. Extensive experiments on 17 medical image classification datasets and 12 corrupted medical image datasets demonstrate that MedViTV2 achieved state-of-the-art results in 27 out of 29 experiments with reduced computational complexity. MedViTV2 is 44\% more computationally efficient than the previous version and significantly enhances accuracy, achieving improvements of 4.6\% on MedMNIST, 5.8\% on NonMNIST, and 13.4\% on the MedMNIST-C benchmark. Code is available at \url{https://github.com/Omid-Nejati/MedViTV2.git}

%In this paper, we introduce MedViTV2, a co-design of Kolmogorov–Arnold Networks and the MedViT family of components, which significantly improves the accuracy of the original MedViT across 16 medical benchmarks, including MedMNIST, Fetal-Planes-DB, CPN X-ray, Kvasir, and PAD-UFES-20. We also present an ablation study to reveal the benefits of our model's key components, demonstrating it as the first robust model specifically designed for corrupted medical images that simulate multi-center studies. Our model sets a new state-of-the-art performance on both clean and robust benchmarks. Code is available at \url{https://github.com/Omid-Nejati/MedViTV2.git}

\end{abstract}

\end{frontmatter}

%\linenumbers

%%%%%%%%% Introduction %%%%%%%%%
\section{Introduction}
Computer-aided diagnosis (CAD) systems have attracted significant research interest in medical image analysis, aiming to assist clinicians in making diagnostic decisions. These systems are applied to various modalities, including X-ray radiography \cite{dai2024unichest}, computed tomography (CT) \cite{lee2025low}, magnetic resonance imaging (MRI) \cite{loizillon2024automatic}, ultrasound \cite{yang2024hierarchical}, and digital pathology \cite{wang2024breast}. The success of deep learning in this domain is partly attributed to the increasing availability of large-scale datasets.  Large datasets with reliable labels are ideal for training deep neural networks.
However, the collection of labeled medical images remains a significant challenge due to data privacy concerns and the labor-intensive nature of expert annotation. %As a result, smaller datasets (sometimes with imbalanced classes) are often easier to obtain from clinical sites.

%Furthermore, imaging practices vary across different centers with some using different equipment, acquiring images through non-expert personnel, or relying on automated labeling methods based on source information \cite{ju2022improving}. For example,  labels for X-ray images can be extracted from corresponding radiology reports \cite{bustos2020padchest, irvin2019chexpert}. These processes can introduce real-world artifacts or common corruptions into the medical images, potentially affecting the quality and accuracy of the resulting diagnoses.

%
%Furthermore, artifacts in medical imaging pose significant challenges to image interpretation and analysis. These artifacts may occur during MRI scans due to lower resolutions or higher acceleration factors, which are intended to shorten scanning times, leading to less detailed images that can compromise diagnoses \cite{eisenstein2024pushing}. Similarly, efforts to minimize radiation exposure in CT and X-ray imaging through reduced dosage can increase noise and reduce image clarity \cite{paprottka2023impact}, thereby affecting the performance of diagnostic algorithms. Additionally, inherent patient anatomical features, such as shadow artifacts in lung ultrasound \cite{vasquez2023automatic}, and the presence of metal implants in MRI \cite{sacher2024role}, which cause streaking artifacts, further complicate image quality. This necessitates the use of advanced deep neural networks to ensure accurate diagnostics.

CAD systems continue to encounter challenges in the medical domain, particularly when dealing with artifacts \cite{eisenstein2024pushing, sacher2024role} and corruptions \cite{huang2023assessing}. These corruptions often arise from various factors, including post-processing techniques, acquisition protocols, data handling, and differences in imaging equipment (e.g., vendor variations). Fortunately, several studies have sought to simulate common corruptions across different medical modalities, including digital pathology \cite{zhang2022benchmarking},  dermatology \cite{maron2021benchmark}, blood microscopy \cite{zhang2020corruption}, and multimodal imaging \cite{di2024medmnist}. While these efforts are foundational, they underscore the need for a robust deep neural network capable of maintaining high performance across diverse medical imaging modalities under such challenging conditions.

\begin{figure*}[!]
    %\vspace{12pt}
    \centering
    \includegraphics[width=0.6\textwidth]{Acc.pdf}
    \caption{Comparison between MedViTs (V1 and V2), MedMamba, and the baseline ResNets, in terms of Average Accuracy vs. FLOPs trade-off over all MedMNIST datasets. MedViTV2-T/S/L significantly improves average accuracy by 2.6\%, 2.5\%, and 4.6\%, respectively, compared to MedViTV1-T/S/L.}
    \label{fig:lineplot}
    \vspace{-1em}
\end{figure*}

To address the aforementioned challenges, Convolutional Neural Networks (CNNs) have made a significant impact in medical imaging by enabling the use of generic feature learning methods across a variety of medical imaging tasks, thanks to their strong ability to learn local representations \cite{xu2024whole, bian2024broadband, zhu2024mp}. In recent years, Vision Transformers (ViTs) \cite{yang2024token, ling2023mtanet, wang2024recursive} have also gained popularity due to their effectiveness in capturing long-range dependencies, especially as model and dataset sizes increase. More recently, MedViTV1 \cite{manzari2023medvit} introduced a hybrid architecture that combines the local feature learning strengths of CNNs with the global feature-capturing capabilities of Transformers, offering a versatile solution for a wide range of medical image datasets, including  MedMNIST~\cite{yang2023medmnist}.

\begin{figure*}[!th]
    \centering
    \includegraphics[width=\textwidth]{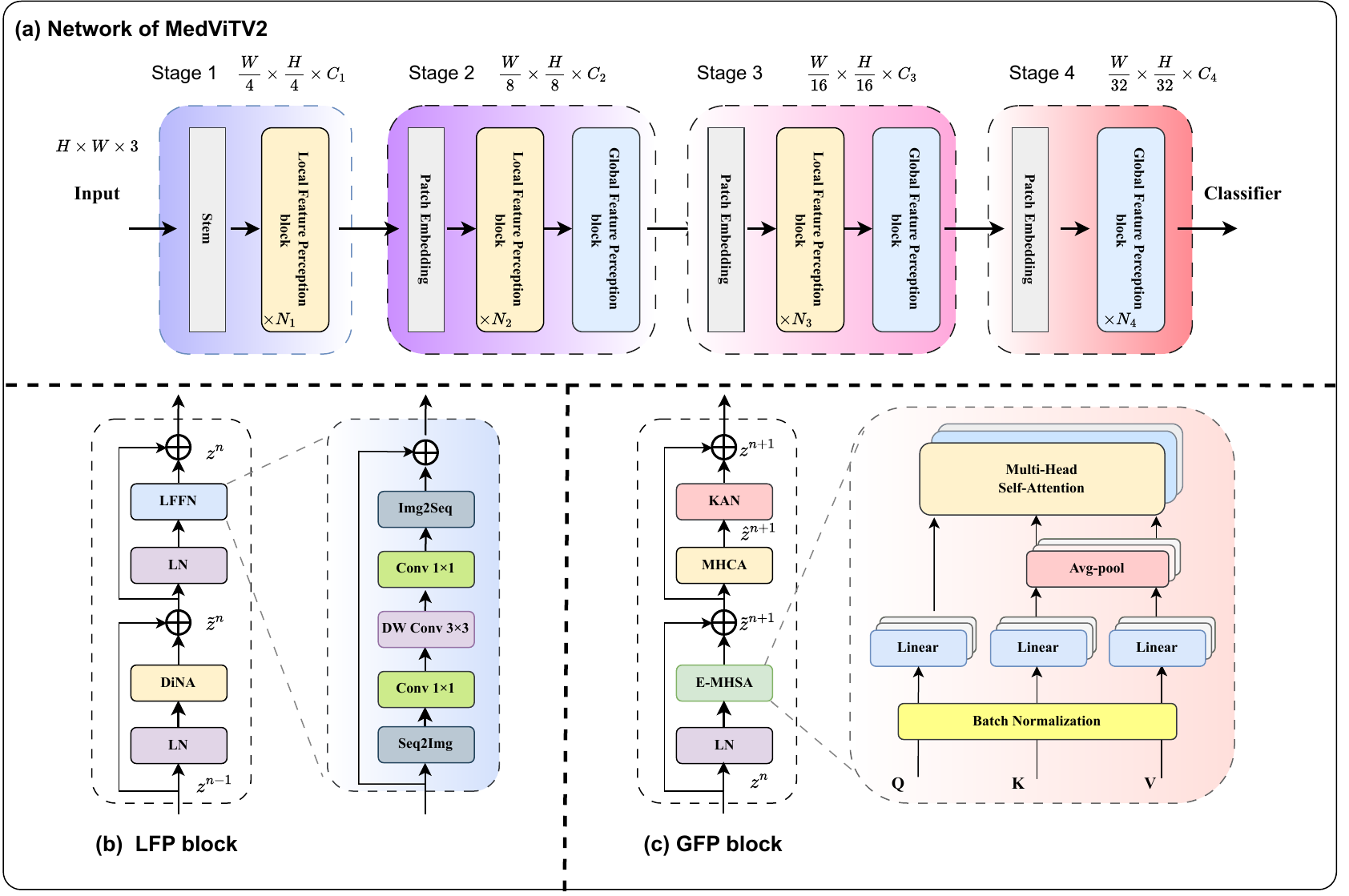}
    \vspace{-4.5mm}
    \caption{Overall architecture of the proposed Medical Vision Transformer (MedViTV2).}
    \label{fig:overall_arch}
    \vspace{-3.5mm}
\end{figure*}

While MedViTV1 demonstrated strong performance on medical image classification benchmarks, it exhibits weaknesses in model scalability, which is instrumental for more complex data and tasks, with accuracy dropping as the number of parameters increases, as shown in Figure \ref{fig:lineplot}. This motivates us to explore components that can support large-scale training, boosting the expressiveness and efficiency of our model while enhancing its competitiveness across a wide range of medical benchmarks. Recently, Dilated Neighborhood Attention (DiNA) \cite{hassani2023neighborhood} introduced an efficient and scalable sliding window attention mechanism in vision tasks. DiNA is a pixel-wise operation that localizes self-attention to the nearest neighboring pixels, enabling linear time and space complexity. The sliding window pattern allows DiNA to capture more global context and exponentially expand receptive fields at no additional cost.
Additionally, Kolmogorov-Arnold Networks (KANs) \cite{liu2024kan} have emerged as a powerful alternative to multi-layer perceptrons (MLPs). In most MLP-based neural networks, functional combinations occur within the activation functions. In contrast, KANs directly perform these combinations on the functions that map inputs to outputs. As a result, KANs rely on simpler functional modules rather than complex, heavily parameterized architectures. This reduction in complexity enables training on medical datasets, which are typically smaller than computer vision datasets. Recent studies have begun to explore the effectiveness of incorporating KAN layers into Transformers, demonstrating that this integration can enhance both the expressiveness and efficiency of Transformers \cite{yang2024kolmogorov, genet2024temporal}, thereby improving their competitiveness across a wide range of applications. Building on this, we incorporate KAN layers into MedViT with the goal of effectively capturing complex non-linear medical structures, while their functional decomposition helps separate and capture the true signal from the noise, which is often prevalent in medical images.

To this end, we propose to leverage the MedViT architecture and the KAN jointly within a hierarchical framework to enable effective large-scale training for the MedViT family and achieve excellent performance on both medical classification tasks and corrupted medical images. Additionally, we conduct a performance analysis of various MedViT component combinations. During this analysis, we identify a potential issue of feature collapse in the Multi-Head Convolutional Attention (MHCA) block when evaluating MedViT on the MedMNIST-C dataset \cite{di2024medmnist}. Therefore, we adapt DiNA to our model to enhance sparse global feature competition. This modification proves most effective when DiNA is combined with the Local Feed-Forward Network (LFFN), suggesting that the optimal design requires complementary components that strengthen robustness and accuracy, and balance global and local feature learning.

In summary, the most significant contributions of our work are:
\begin{itemize}
    \item To the best of our knowledge, this is the first study that integrates KAN into the feed-forward pathway of Transformers for medical image classification. This adaptation reduces computational complexity by 44\% while significantly enhancing performance.
    \item We propose the DiNA block, an efficient and powerful sparse global attention mechanism. This innovation enables our model to scale up and address the feature collapse issue present in previous versions of our MedViT family.
    \item We introduce a novel Hierarchical Hybrid Strategy that meticulously designs our model to balance global and local feature perception. This strategy boosts performance with high efficiency.
    \item Our extensive experiments across 17 medical image datasets and 12 corrupted medical image benchmarks demonstrate that MedViTV2s (MedViTV2-tiny, MedViTV2-small, MedViTV2-base, and MedViTV2-large) achieve state-of-the-art performance on most of them.
\end{itemize}
%Our code is available at \url{https://github.com/Omid-Nejati/MedViTV2.git}

%In summary, we introduce the Medical Vision Transformer V2 (MedViTV2), a model specifically designed to address scaling challenges observed in prior versions. By integrating DiNA and KAN within a hierarchical framework, MedViTV2 achieves consistent performance improvements across diverse medical datasets and benchmarks, including natural and corrupted data. Our findings show that this model achieves state-of-the-art (SOTA) performance on several datasets, including MedMNIST~\cite{yang2023medmnist}, MedMNIST-C~\cite{yang2023medmnist}, Fetal-Planes-DB~\cite{burgos2020evaluation}, CPN X-ray~\cite{kumar2022litecovidnet}, PAD-UFES-20~\cite{pacheco2020pad}, and Kvasir~\cite{pogorelov2017kvasir}. %To the best of our knowledge, this is the first research to design a robust model for corrupted medical image classification.

%
%

\section{Related Works}
\label{sec:2}

\subsection{Medical Image Classification.}
Medical image classification remains a significant challenge, critical in organizing large volumes of data into meaningful categories \cite{azad2024advances}. Over the past decade, CNNs have dominated the field of image classification. They have been widely employed in applications such as cancer detection \cite{ren2023ipsilateral}, skin disease diagnosis \cite{dai2024i2u, dai2022ms}, thoracic surgery \cite{harirpoush2024architecture}, retinal disease identification \cite{ju2023hierarchical}, and fetal brain volume estimation \cite{pei2023pets}.

More recently, Vision Transformers (ViTs) have emerged as a powerful alternative to conventional CNNs, achieving remarkable success in various image classification tasks \cite{liu2021swin, dai2021transmed, chu2021twins, CSWin}. ViTs offer several advantages, including the ability to model long-range dependencies, adapt to diverse inputs, and generate attention maps that highlight critical regions within an image \cite{gheflati2022vision}. These features have sparked significant interest in leveraging Transformer-based models for medical image classification, where precise classification is increasingly essential to support timely clinical decision-making, particularly for difficult cases.
Early ViT models typically rely on large-scale datasets and relatively simple configurations \cite{ViT}. However, recent advancements have integrated inductive biases related to visual perception into ViT architectures \cite{manzari2023robust, yan20233d, wang2024repvit, yan2023hybrid}. This evolution has made ViTs more adaptable and effective in classification, registration, and segmentation. By treating images as sequences of patches without incorporating 2D inductive biases, ViTs are particularly suitable for multimodal applications \cite{xu2023research, hatamizadeh2021swin}.
In particular, the growth of datasets and innovations in model architectures \cite{xue2024neural, xue2024gradient, jiang2025score} have driven ViT-based foundation models with unprecedented capabilities, enabling \textit{flexible} applications in medical imaging \cite{manigrasso2025mammography, horst2024cellvit, koleilat2024medclip}. For example, researchers have introduced FastGlioma, a tool for detecting tumor infiltration during surgery \cite{kondepudi2024foundation}, while RETFound learns generalizable representations from unlabelled retinal images, enabling label-efficient model adaptation across various applications \cite{zhou2023foundation}. Additional applications include leveraging tumor registry data and demographic information to predict overall survival rates \cite{jee2024automated}. %With further optimization and domain-specific enhancements, ViTs are well-positioned to play an increasingly transformative role in medical image analysis.

\subsection{Kolmogorov–Arnold Networks}
KANs have inspired numerous studies that demonstrate their effectiveness across various domains, including keyword spotting, complex optimization problems, survival analysis, time series forecasting, quantum computing, and vision tasks~\cite{li2025u,somvanshi2024survey,DBLP}. Furthermore, many advanced KAN models leverage well-established mathematical functions, particularly those adept at handling curves, such as B-splines~\cite{ta2024bsrbf}, which combine B-splines and Radial Basis Functions (RBF) to fit input data during training. FastKAN~\cite{li2024kolmogorov} approximates third-order B-splines in KANs using Gaussian RBF, while DeepOKAN~\cite{abueidda2024deepokan} directly employs Gaussian RBF instead of B-splines. Other approaches, such as FourierKAN-GCF~\cite{xu2024fourierkan}, wavelet-based KANs~\cite{seydi2024unveiling}, and polynomial-function-based KANs~\cite{teymoor2024exploring}, explore the suitability of various basis functions in KAN models for classification and other tasks.

KANs have also demonstrated significant potential in medical image processing, where interpretability and precision are paramount. For instance, BSRBF-KAN~\cite{ta2024bsrbf} has been utilized to improve the segmentation accuracy of complex medical images, such as MRI scans, by leveraging the flexibility of RBF. Similarly, TransUKAN~\cite{wu2024transukan} integrates KANs, Transformers, and U-Net architectures to enhance the efficiency and performance of medical image segmentation while significantly reducing parameter counts. Additionally, Bayesian-KAN~\cite{hassan2024bayesian} combines KAN with Bayesian inference to deliver explainable, uncertainty-aware predictions in healthcare settings. These developments showcase how the mathematical foundations of KANs can be adapted to address specific challenges in medical imaging, providing a balance between high interpretability and computational efficiency.
%This growing body of evidence underscores the versatility of KANs and their applicability across diverse fields, including healthcare, establishing them as a powerful tool for advancing AI-driven solutions in critical domains.

\section{Methods}

In this section, we first give a brief overview of the proposed MedViT. Then, we describe the main body designs within MedViT-V2, which include the Local Feature Perception (LFP), Global Feature Perception (GFP), and hierarchical hybrid strategy. In addition, we provide different model sizes for the proposed architecture.

\subsection{Overview of MedViTV1}

MedViT is a general model for medical image classification that has achieved excellent performance across a wide range of medical datasets, including chest X-rays, diabetic retinopathy, and various body organs. The core idea of MedViT is to incorporate the locality of CNNs into different components of Transformers, such as the feed-forward network and multi-head attention, thereby combining the strengths of both CNNs and Transformers. CNN blocks have a strong intrinsic ability to capture high-frequency features, while Transformers excel at extracting low-frequency features. As a result, MedViT can combine these rich features, leading to greater accuracy than pure CNNs and Transformers.

\subsection{Dilated Attention Block}

To introduce an efficient attention into LFP block of our model at no additional cost, we use dilated attention as shown in Figure \ref{fig:overall_arch}. Dilated attention confines self-attention to the nearest neighbors of each pixel, maintaining the same network complexity as well as parameter count as shifted windows attention \cite{liu2021swin}. Nonetheless, it operates within overlapping shifted windows, thereby preserving translation equivariance. Specifically, let $\sqrt{d}$ denote a scaling parameter and $d$ represent the embedding dimension. In DiNA, for a given dilation value $\delta$, we determine $\rho_j^\delta(i)$ as the $j^{th}$ nearest neighbor of token $i$ that satisfies the condition  $i \bmod \delta = j \bmod \delta$. Using this definition, the $\delta$-dilated neighborhood attention weights for the $i^{th}$ token, with a neighborhood size $k$, $\mathbf{A}_i^{(k, \delta)}$, can be expressed as follows:

$$
\mathbf{A}_i^{(k, \delta)}=\left[\begin{array}{c}
Q_i K_{\rho_1^\delta(i)}^T+B_{\left(i, \rho_1^\delta(i)\right)} \\
Q_i K_{\rho_2^\delta(i)}^T+B_{\left(i, \rho_2^\delta(i)\right)} \\
\vdots \\
Q_i K_{\rho_k^\delta(i)}^T+B_{\left(i, \rho_k^\delta(i)\right)}
\end{array}\right]
\vspace{1em}
$$

\noindent
where query ($Q$) and key ($K$) are linear projections of the input data, while $B(i,j)$ represents the relative bias between token $i$ and token $j$. Similarly, $\mathbf{V}_i^{(k, \delta)}$ is determined as $\delta$-dilated adjacent values for the $i^{th}$ token, incorporating $k$ neighboring tokens:

$$
\mathbf{V}_i^{(k, \delta)}=\left[\begin{array}{llll}
V_{\rho_1^\delta(i)}^T & V_{\rho_2^\delta(i)}^T & \ldots & V_{\rho_k^\delta(i)}^T
\end{array}\right]^T
\vspace{1em}
$$

\noindent
Next, the output of DiNA for $i^{th}$ token is formulated as follows:

$$
\operatorname{DiNA}_k^\delta(i)=\operatorname{softmax}\left(\frac{\mathbf{A}_i^{(k, \delta)}}{\sqrt{d_k}}\right) \mathbf{V}_i^{(k, \delta)}
$$

\noindent
The LFP block is a collaborative operation between DiNA and LFFN to capture both local and global features within the input data. The mathematical formulation is as follows:

\begin{equation}
\tilde{z}^{n} =\operatorname{DiNA}(LN(z^{n-1}))+z^{n-1},
\end{equation}

\begin{equation}
z^{n} =\operatorname{LFFN}(LN(\tilde{z}^{n}))+\tilde{z}^{n}.
\end{equation}

\noindent
In the provided equations, $z^{n-1}$ undergoes layer normalization (LN) before entering the $DiNA$ module. Also, $\tilde{z}^{n}$ and $z^{n}$ denote the output of $DiNA$ and $LFFN$ for the $n^{th}$ block of LFP.

\subsection{Kolmogorov–Arnold Networks (KANs)}

We incorporate KANs into the GFP block of our model architecture. The exceptional efficiency and interpretability of KANs, as outlined in \cite{liu2024kan}, form the foundation of this strategy.
One way to describe a $N$-layer KAN is as a composition of multiple KAN layers arranged sequentially:
\begin{equation}
    \operatorname{KAN}(\mathbf{Z})=\left(\boldsymbol{\Phi}_{N-1} \circ \boldsymbol{\Phi}_{N-2} \circ \cdots \circ \boldsymbol{\Phi}_{1} \circ \boldsymbol{\Phi}_{0}\right) \mathbf{Z},
\end{equation}
where $\boldsymbol{\Phi}_i$ signifies the KAN network's $i$-th layer. $\boldsymbol{\Phi}$ consists of $ m_{in} \times m_{out}$ learnable activation functions $\phi$ for each KAN layer, which has $m_{in}$ -dimensional input and $m_{out}$ -dimensional output:
\begin{equation}
    \boldsymbol{\Phi}=\left\{\phi_{p, q}\right\}, \quad q=1,2, \cdots, m_{\text {in }}, \quad p=1,2 \cdots, m_{\text {out }},
\end{equation}

The computation of the KAN network from layer $n$ to layer $n+1$ can be demonstrated in matrix form as $\mathbf{Z}{n+1} = \boldsymbol{\Phi}{n}\mathbf{Z}_{n}$, where:
\begin{equation}
\boldsymbol{\Phi}_{n} = \left(\begin{array}{cccc}
\phi_{n, 1,1}(\cdot) & \phi_{n, 1,2}(\cdot) & \cdots & \phi_{n, 1, m_{n}}(\cdot) \\
\phi_{n, 2,1}(\cdot) & \phi_{n, 2,2}(\cdot) & \cdots & \phi_{n, 2, m_{n}}(\cdot) \\
\vdots & \vdots & & \vdots \\
\phi_{n, m_{n+1}, 1}(\cdot) & \phi_{n, m_{n+1}, 2}(\cdot) & \cdots & \phi_{n, m_{n+1}, m_{n}}(\cdot)
\end{array}\right)
\vspace{1em}
\end{equation}
KANs differ from traditional MLPs by eliminating the need for linear weight matrices. Instead, they employ parameterized functions as weights and integrate learnable activation functions along the edges. This architectural design enables KANs to achieve superior performance while reducing model size.

The first KAN was implemented using a function, $\phi(x)$, defined as the sum of a spline function and a base function, each associated with their respective weight matrixes, $w_s$ and $w_b$:

\begin{equation}
\begin{aligned}
\phi(x) = w_b b(x) + w_s spline(x)
\end{aligned}
\label{eq:acti_funct_imp}
\end{equation}

\begin{equation}
\begin{aligned}
b(x) = silu(x) = \frac{x}{1 + e^{-x}}
\end{aligned}
\label{eq:b_function}
\end{equation}

\begin{equation}
\begin{aligned}
spline(x) = \sum_{i}c_iB_i(x)
\end{aligned}
\label{eq:spline_function}
\vspace{1em}
\end{equation}
where \(b(x)\) equals to $silu(x)$ as in Equation~\ref{eq:b_function} and \(spline(x)\) is defined as a linear combining of B-splines $B_i$s and control points (coefficients) $c_i$s as in Equation~\ref{eq:spline_function}. Each activation function is triggered with \(spline(x) \approx 0\) and \(w_s = 1\) , while \(w_b\) is activated by utilizing Xavier initialization.

Building upon the approach introduced in FastKAN~\cite{li2024kolmogorov}, which enhances training efficiency by leveraging Gaussian radial basis functions (RBFs) to approximate 3rd-order B-splines and incorporating layer normalization to maintain input values within the RBF domain, we adopt Reflectional Switch Activation Functions (RSWAFs). RSWAFs, as a variant of RBFs, are computationally efficient due to their homogeneous grid pattern. The RSWAF is defined as:

\begin{equation}
\begin{aligned}
\phi_{\mathit{RSWAF}}(r) = 1 - \left(\tanh\left(\frac{r}{h}\right)\right)^2
\end{aligned}
\label{eq:rswaf_funct}
\end{equation}

\begin{equation}
\begin{aligned}
\mathit{RSWAF}(x) &= \sum_{i=1}^{N} w_i \phi_{\mathit{RSWAF}}(r_i) \\
&= \sum_{i=1}^{N} w_i \left(1 - \left(\tanh\left(\frac{||x - c_i||}{h}\right)\right)^2\right)
\end{aligned}
\label{eq:rswaf_network}
\vspace{1em}
\end{equation}

\noindent
The complete implementation of the GFP can be derived as outlined below:

\begin{equation}
    \begin{aligned}
        & \tilde{z}^{n+1} =\mathrm{E}-\mathrm{MHSA}(LN(z^{n}))+z^{n}, \\
        & \hat{z}^{n+1} =\operatorname{MHCA}(\tilde{z}^{n+1}), \\
        & \bar{z}^{n+1} =\operatorname{KAN}(\hat{z}^{n+1}),\\
        & z^{n+1} =\operatorname{Concat}(\tilde{z}^{n+1}, \bar{z}^{n+1}), \\
    \end{aligned}
\vspace{1em}
\end{equation}

\noindent
where $\tilde{z}^{n+1}$, $\hat{z}^{n+1}$, $\bar{z}^{n+1}$, and $z^{n+1}$ are the output of $E-MHSA$, $MHCA$, $KAN$, and GFP, respectively (see Figure~\ref{fig:overall_arch}). Additionally, Layer Normalization (LN) and ReLU are uniformly used in GFP as efficient normalization and activation functions. Compared to MedViTV1, GFP is capable of capturing and scaling rich features in a lightweight and robust manner.

\begin{table}[]
    \centering\caption{Comparison of various architecture designs. `Cls' represents the accuracy achieved on the TissueMNIST dataset, while throughput is consistently measured for an input size of $3 \times 224 \times 224$. `T' and `C' indicate Transformer and convolution blocks, respectively, and `H' refers to our hierarchical hybrid structure. }
    \resizebox{0.65\textwidth}{!}{
        \begin{tabular}{c|cc|c|c}
            \toprule
            \multirow{2.5}{*}{Model} & \multicolumn{2}{c|}{Complexity}  & Latency         & Cls      \\ \cmidrule(l){2-5}
                                     & Param(M)  & FLOPs (G)           & throughput(ms)     & Acc(\%)   \\ \midrule
            C C C C                                 & 27.4               & 7.2          &  23.3                          & 68.76           \\
            C C C T                                 & 28.5               & 7.4          & 23.5                          & 68.93          \\
            C C T T                                 & 30.7               & 7.8         & 24.6                          & 68.37          \\
            C T T T                                 & 31.0               & 7.8          & 24.9                          & 68.28          \\ \midrule
            C C C H$_\text{N}$                                     & 27.1               & 6.9          & 18.8          & 68.80          \\
            C C H$_\text{N}$ H$_\text{N}$                         & 28.6               & 7.4          & 19.4              & 69.58          \\
            C H$_\text{N}$ H$_\text{N}$ H$_\text{N}$       & 29.3  & 7.6  & 19.6      & 69.73          \\
            \textbf{C H$_\text{N}$ H$_\text{N}$ T}      & \textbf{29.6}               & \textbf{7.6} & \textbf{20.1}      & \textbf{70.51}          \\
            \bottomrule
    \end{tabular}}
    \label{tab:NHS1}
\end{table}

\begin{table*}[]
    \centering\caption{Comparative Analysis of Patterns and Hyperparameter Configurations within the Hierarchical Hybrid Strategy (HHS). S1, S2, S3, and S4 denote Stage 1, Stage 2, Stage 3, and Stage 4, respectively.}
    \resizebox{0.68\textwidth}{!}{
        \begin{tabular}{c|c|c|c|c|c|c|c|c|c}
            \toprule
            \multirow{2.5}{*}{Hybrid Strategy}       &\multirow{2.5}{*}{Model}            & \multicolumn{4}{c|}{Configs}              &  \multicolumn{2}{c|}{Complexity}    & \multirow{2.5}{*}{Latency(ms)}   & \multirow{2.5}{*}{Acc(\%)}       \\ \cmidrule(l){3-8}
                                                     &                                    & S1 & S2 & S3 & S4 &  {param(M)} &   {FLOPs(G)}   &   &     \\ \midrule
            \multirow{3}{*}{$\begin{bmatrix} \text{LFP} \times N \\ \text{GFP} \times 1 \end{bmatrix}$}
            & \multirow{3}{*}{Small}   & 2 & 2 & (8+1) & 2             & 27.0           & 6.9          & 20.4                          & 68.35          \\
            &     & 2 & 2 & (11+1) & 2             & 29.1           & 7.7         & 20.9                          & 69.59          \\
            &    & 2 & 2 & (14+1) & 2             & 31.1           & 8.5          & 21.6                          & 69.80          \\ \midrule
            \multirow{5}{*}{$\begin{bmatrix} \text{LFP} \times N \\ \text{GFP} \times 1 \end{bmatrix} \times L$}    & \multirow{5}{*}{Small}   & 2 & 2 & (1+1)$\times$3 & 2            & 31.5           & 8.3          & 20.7                          & 69.31          \\
                        &                           & 2 & 2 & (2+1)$\times$3 & 2       & 33.5           & 9.1          & 21.2                         & 70.54          \\
                        &                           & 2 & 2 & (3+1)$\times$3 & 2       & 35.5           & 9.8          & 21.5                          & 70.54          \\
                        &                           & \textbf{2} & \textbf{2} & \textbf{(2+1)$\times$2} & \textbf{2}       &\textbf{29.6}   &\textbf{7.6}  & \textbf{20.1}        & \textbf{70.51}          \\
                        &                           & 2 & 2 & (3+1)$\times$2 & 2       & 30.9           & 8.2          & 20.7                          & 70.52          \\
            \bottomrule
    \end{tabular}}
    \label{tab:NHS2}
\vspace{-0.25cm}
\end{table*}

\begin{table*}[]
  \centering
  \caption{Detailed configurations of MedViTV2 variants. $C$ and $S$ denote the number of channels and stride of convolution for each stage.}
  \label{tab:configs}
  \resizebox{0.85\textwidth}{!}{
  \begin{tabular}{ccc|c|c|c|c}
  \toprule
  Stages                        & Output size                                         & Layers                                                          & MedViT-T                & MedViT-S            & MedViT-B  & MedViT-L       \\  \midrule
  \multirow{5}{*}{Stem}         & \multirow{5}{*}{$\displaystyle{\frac{H}{4}} \times \frac{W}{4} $}  & \multirow{5}{*}{\shortstack{Convolution \\ Layers}}             & \multicolumn{4}{c}{$\text{Conv}\ 3\times3, C=64, S=2$}                       \\  \cmidrule(l){4-7}
                                &                                                     &                                                                 & \multicolumn{4}{c}{$\text{Conv}\ 3\times3, C=32, S=1$}                       \\  \cmidrule(l){4-7}
                                &                                                     &                                                                 & \multicolumn{4}{c}{$\text{Conv}\ 3\times3, C=64, S=1$}                       \\  \cmidrule(l){4-7}
                                &                                                     &                                                                 & \multicolumn{4}{c}{$\text{Conv}\ 3\times3, C=64, S=2$}                       \\  \midrule
  \multirow{4}{*}{Stage 1}      & \multirow{4}{*}{$\displaystyle{\frac{H}{4}} \times \frac{W}{4} $}  & \multirow{1}{*}{Patch Embedding}                                & \multicolumn{4}{c}{$\text{Conv}\ 1\times1, C=96$}                            \\  \cmidrule(l){3-7}
                                &                                                     & \multirow{2}{*}{\shortstack{MedViT \\ Block}}            & \multicolumn{4}{c}{\multirow{2}{*}{$\begin{bmatrix} \text{LFP} \times 2, 96 \end{bmatrix} \times 1$}}                    \\
                                &                                                     &                                                                 & \multicolumn{4}{c}{}                                                                                          \\  \midrule
  \multirow{6}{*}{Stage 2}      & \multirow{6}{*}{$\displaystyle{\frac{H}{8}} \times \frac{W}{8} $}  & \multirow{2.5}{*}{Patch Embedding}                              & \multicolumn{4}{c}{$\text{Avg\_pool}, S=2$}                                                                          \\  \cmidrule(l){4-7}
                                &                                                     &                                                                 &  $C=128$ & $C=128$ & $C=192$ & $C=256$                                                                   \\  \cmidrule(l){3-7}
                                &                                                     & \multirow{3}{*}{\shortstack{MedViT \\ Block}}            &
                                \multirow{3}{*}{$\begin{bmatrix} \text{LFP} \times 1, 128 \\ \text{GFP} \times 1, 128 \end{bmatrix} \times 1$}        & \multirow{3}{*}{$\begin{bmatrix} \text{LFP} \times 1, 128 \\ \text{GFP} \times 1, 128 \end{bmatrix} \times 1$}        & \multirow{3}{*}{$\begin{bmatrix} \text{LFP} \times 1, 192 \\ \text{GFP} \times 1, 192 \end{bmatrix} \times 1$} & \multirow{3}{*}{$\begin{bmatrix} \text{LFP} \times 1, 256 \\ \text{GFP} \times 1, 256 \end{bmatrix} \times 1$}       \\
                                &                                                     &                                                                 &                                                                                               &                                                                                               &                                                                                              \\
                                &                                                     &                                                                 &                                                                                               &                                                                                               &
                                 \\  \midrule
  \multirow{6}{*}{Stage 3}      & \multirow{6}{*}{$\displaystyle{\frac{H}{16}} \times \frac{W}{16}$} & \multirow{2.5}{*}{Patch Embedding}                              & \multicolumn{4}{c}{$\text{Avg\_pool}, S=2$}                                                                          \\  \cmidrule(l){4-7}
                                &                                                     &                                                                 & $C=192$ & $C=256$ & $C=384$ & $C=512$                                                                   \\  \cmidrule(l){3-7}
                                &                                                     & \multirow{3}{*}{\shortstack{MedViT \\ Block}}            & \multirow{3}{*}{$\begin{bmatrix} \text{LFP} \times 2, 192 \\ \text{GFP} \times 1, 192 \end{bmatrix} \times 2$}        & \multirow{3}{*}{$\begin{bmatrix} \text{LFP} \times 2, 256 \\ \text{GFP} \times 1, 256 \end{bmatrix} \times 2$}        & \multirow{3}{*}{$\begin{bmatrix} \text{LFP} \times 2, 384 \\ \text{GFP} \times 1, 384 \end{bmatrix} \times 2$} & \multirow{3}{*}{$\begin{bmatrix} \text{LFP} \times 2, 512 \\ \text{GFP} \times 1, 512 \end{bmatrix} \times 2$}       \\
                                &                                                     &                                                                 &                                                                                               &                                                                                               &                                                                                              \\
                                &                                                     &                                                                 &                                                                                               &                                                                                               &                                                                                              \\  \midrule
  \multirow{4}{*}{Stage 4}      & \multirow{4}{*}{$\displaystyle{\frac{H}{32}} \times \frac{W}{32}$} & \multirow{2.5}{*}{Patch Embedding}                              & \multicolumn{4}{c}{$\text{Avg\_pool}, S=2$}                                                                          \\  \cmidrule(l){4-7}
                                &                                                     &                                                                 & $C=384$ & $C=512$ & $C=768$ & $C=1024$                                                                  \\  \cmidrule(l){3-7}
                                &                                                     & \multirow{2}{*}{\shortstack{MedViT \\ Block}}            & \multirow{2}{*}{$\begin{bmatrix} \text{GFP} \times 1, 384 \end{bmatrix} \times 1$}        & \multirow{2}{*}{$\begin{bmatrix} \text{GFP} \times 1, 512 \end{bmatrix} \times 2$}        & \multirow{2}{*}{$\begin{bmatrix}  \text{GFP} \times 1, 768 \end{bmatrix} \times 2$} & \multirow{2}{*}{$\begin{bmatrix} \text{GFP} \times 1, 1024 \end{bmatrix} \times 2$}       \\
                                &                                                     &                                                                 &                                                                                               &                                                                                               &                                                                                              \\  \bottomrule
  \end{tabular}
}
\vspace{-0.25cm}
\end{table*}

\begin{algorithm}[h]
\footnotesize
\caption{MedViTV2: KAN-Integrated Transformers with Dilated Neighborhood Attention}
\label{alg:medvitv2}
\DontPrintSemicolon
\SetKwInOut{Input}{Input}\SetKwInOut{Output}{Output}\SetKwInOut{Params}{Params}
\Input{Image $x\in\mathbb{R}^{H\times W\times 3}$}
\Params{Stage widths $\{C_i\}_{i=1}^{4}$, stage repeats $\{L_i\}_{i=1}^{4}$, LFP repeats per group $\{N_i\}_{i=1}^{4}$;\;
E-MHSA spatial reduction ratios per stage (e.g., $[8,4,2,1]$);\;
DiNA neighborhood size $k$ and dilation $\delta$;\;
Classifier weights $W_{\text{cls}}$.}
\Output{Class probabilities $\hat{y}$}
\BlankLine
\textbf{Stem (overall stride 4).} $z_0 \leftarrow \texttt{ConvStem}(x)$ \tcp*{stack of small convs/downsampling}
\BlankLine
\For{$i \leftarrow 1$ \KwTo $4$}{
  \textbf{Patch Embedding / Downsample to Stage $i$.} $z \leftarrow \texttt{PatchEmbed}_i(z_{i-1})$ \tcp*{e.g., AvgPool (stride 2) + $1{\times}1$ conv}
  \For{$\ell \leftarrow 1$ \KwTo $L_i$}{
    \For{$n \leftarrow 1$ \KwTo $N_i$}{
      \textbf{Local Feature Perception (LFP).}\;
      \Indp
      $u \leftarrow \texttt{LN}(z)$\;
      $u \leftarrow \texttt{DiNA}(u;\,k,\delta)$ \tcp*{dilated neighborhood attention}
      $z \leftarrow z + u$ \tcp*{residual}
      $v \leftarrow \texttt{LN}(z)$\;
      $v \leftarrow \texttt{LFFN}(v)$ \tcp*{$1{\times}1$ conv $\rightarrow$ DW-$3{\times}3$ conv $\rightarrow$ $1{\times}1$ conv, with BN+ReLU}
      $z \leftarrow z + v$ \tcp*{residual}
      \Indm
    }
    \If{$i > 1$}{
      \textbf{Global Feature Perception (GFP).}\;
      \Indp
      $q \leftarrow \texttt{LN}(z)$\;
      $\tilde{z} \leftarrow z + \texttt{E\,-\,MHSA}(q;\,\text{reduction ratio at stage }i)$ \tcp*{efficient MHSA + residual}
      $h \leftarrow \texttt{MHCA}(\tilde{z})$ \tcp*{multi-head convolutional attention}
      $g \leftarrow \texttt{KAN}(h)$ \tcp*{KAN feed-forward (e.g., expansion then projection)}
      $z \leftarrow \texttt{Concat}(\tilde{z},\, g)$ \tcp*{channel concat of residual branch and KAN output}
      \Indm
    }
  }
  $z_i \leftarrow z$
}
\BlankLine
\textbf{Head.} $p \leftarrow \texttt{GAP}(z_4)$ \tcp*{global average pooling}
\quad $\hat{y} \leftarrow \texttt{Softmax}(W_{\text{cls}}\, p)$\;
\Return $\hat{y}$
\end{algorithm}

\subsection{Hierarchical Hybrid Strategy}\label{sec:strategy}

In traditional hybrid models, convolutional layers are commonly used in the initial stages of hybrid architectures, with Transformer blocks typically stacked toward the network's end. However, this conventional approach may struggle to capture global information effectively in the early layers, potentially leading to subpar performance. To address this, we propose a novel hierarchical hybrid strategy, which is delineated in bold in Table \ref{tab:NHS1}. In this table, `T' represents the uniform incorporation of a Transformer stage (GFP), while `C' denotes the consistent layering of convolution blocks (LFP). The $H_N$ adopts an $(LFP\times N+GFP \times1)$ pattern, where each stage comprises one GFP block and N times LFP blocks, except the first stage, which does not have a GFP block. The detailed configuration of MedViT-V2 architecture is listed in Table \ref{tab:configs}. By explicitly incorporating a GFP block at the end of each stage, this design allows the model to learn global features effectively, even in the shallow layers. Furthermore, each stage is repeated $L$ times, resulting in the final model pattern of $(LFP\times N + GFP\times 1) \times L$. This iterative structure enhances the model's capacity to extract and integrate local and global information across multiple stages, ultimately improving its performance in capturing complex patterns and relationships within the data.

To further assess the generalized efficacy of $\text{C}\,\text{H}_\text{N}\,\text{H}_\text{N}\,\text{T}$, we scaled up a MedViT model by augmenting the block count in its third stage. The initial three rows of experimental results in Table~\ref{tab:NHS2} reveal that a mere increase in the number of blocks yields diminishing performance returns, eventually leading to saturation.
This observation implies that a straightforward expansion of model size by simply increasing the value of $N$ in the $(\text{LFP} \times N + \text{GFP} \times 1)$ pattern, i.e., adding more convolution blocks, is not the most effective approach.
Prompted by this, we conducted extensive experiments to meticulously investigate the impact of $N$ on model performance. As presented in the middle section of Table~\ref{tab:NHS2}, we designed models with diverse $N$ configurations for the third stage. To facilitate a fair comparison based on similar latency profiles, we employed $L$ stacked groups of the $(\text{LFP} \times N + \text{GFP} \times 1)$ pattern when $N$ was small.
Remarkably, we discovered that models employing the stacked pattern $(\text{LFP} \times N + \text{GFP} \times 1)\times L$ achieved superior performance compared to those using only a single $(\text{LFP} \times N + \text{GFP} \times 1)$ pattern. This suggests that the iterative and principled combination of low-frequency signal extractors (LFP) and high-frequency signal extractors (GFP) through the $(\text{LFP} \times N + \text{GFP} \times 1)$ sub-pattern is crucial for learning higher-quality representations. From Table~\ref{tab:NHS2}, the model with $N=2$ in the third stage demonstrates the optimal balance between performance and latency.
Subsequently, we constructed even larger models by increasing the value of $L$ within the $(\text{LFP} \times 2 + \text{GFP} \times 1)\times L$ pattern for the third stage. As indicated in Table~\ref{tab:NHS2}, for small size of our model, setting $L=3$ and $N=3$ yielded the best results. However, the performance improvement over $L=2$ and $N=2$ was marginal, while simultaneously introducing a significant increase in complexity. This observation, combined with the broader analysis, confirms the wide applicability and effectiveness of our proposed $(\text{LFP} \times N + \text{GFP} \times 1)\times L$ pattern. Consequently, we adopted $N=2$ as the default configuration for the remainder of this paper due to its favorable performance-complexity trade-off.

Based on the aforementioned Hierarchical Hybrid Strategy, we constructed MedViT, formally defined as:
\begin{align}
    \label{equ:nhs}
    \text{MedViT}(X) = \oint_{i=1}^{4} \left\{ \left[ \varGamma_i \left( \varPsi_i (X) \times N_i \right) \right] \times L_i \right\}
\end{align}
where $i \in \{1, 2, 3, 4\}$ denotes the stage index. $\varPsi_i$ refers to the LFP at stage $i$. $\varGamma_i$ represents an identity layer for the first stage ($i=1$) and the GFP for subsequent stages ($i>1$). The operator $\oint$ signifies the sequential concatenation of these stages.

\subsection{MedViTV2 Architectures}
The overall architecture of MedViTV2 is already shown in Figure \ref{fig:overall_arch}, while the detailed procedure is presented in Algorithm \ref{alg:medvitv2}.
To ensure a fair comparison with existing SOTA networks in medical domain, we introduce four representative variants, MedViT-V2-T/S/B/L. The architectural specifications for these variants are detailed in Table \ref{tab:configs}, where $S$ represents the stride of each stage and $C$ denotes the output channel.
The spatial reduction ratio in GFP is set to [8, 4, 2, 1], while the channel shrink ratio $r$ is consistently fixed at 0.75.
The expansion ratios are set to 2 for the KAN layer and and 3 for the feature expansion in LFFN. The head dimension in MHCA and E-MHSA is fixed at 32. Both LFP and GFP employ ReLU as the activation function and BatchNorm as the normalization layer.

\begin{table*}[!ht]
%\scriptsize
\centering
\caption{The detailed descriptions for 17 datasets used in the work. Some of the notations used in datasets include BC: Binary-Class. OR: Ordinal Regression. ML: Multi-Label. MC: Multi-Class.}
%\vspace{-0.7mm}
\label{tab:dataset}
\resizebox{\textwidth}{!}{
\begin{tabular}{lcccccc}
\toprule
Name  & Data Modality & Task (\# Classes / Labels) & \# Samples & \# Training / Validation / Test \\
\midrule
PAD-UFES-20 & Human Skin Smartphone Image & MC (6) & 2,298 & 1,384 / 227 / 687   \\
CPN X-ray &  Chet X-ray & MC (3) & 5,228 & 3,140 / 521 / 1,567 \\
Fetal-Planes-DB &  Maternal-fetal Ultrasound & MC (6) & 1,2400 & 7,446 / 1,237 / 3,717 \\
Kvasir &  Gastrointestinal Endoscope & MC (8) & 4,000 & 2,408 / 392 / 1,200 \\
ISIC2018 & Skin Lesion  & MC (7) & 11,720 & 10,015 / 193 / 1512 \\
ChestMNIST & Chest X-Ray & ML (14) BC (2) & 112,120 & 78,468 / 11,219 / 22,433   \\
PathMNIST &  Colon Pathology & MC (9) & 107,180 & 89,996 / 10,004 / 7,180 \\
OCTMNIST & Retinal OCT & MC (4) & 109,309 & 97,477 / 10,832 / 1,000  \\
DermaMNIST & Dermatoscope & MC (7) & 10,015 & 7,007 / 1,003 / 2,005 \\
RetinaMNIST & Fundus Camera & OR (5) & 1,600 & 1,080 / 120 / 400 \\
PneumoniaMNIST & Chest X-Ray & BC (2) & 5,856 &  4,708 / 524 / 624 \\
BreastMNIST & Breast Ultrasound & BC (2) & 780 & 546 / 78 / 156  \\
TissueMNIST & Kidney Cortex Microscope & MC (8) & 236,386 & 165,466 / 23,640 / 47,280  \\
BloodMNIST & Blood Cell Microscope & MC (8) & 17,092 & 11,959 / 1,712 / 3,421  \\
OrganAMNIST & Abdominal CT & MC (11) & 58,850 & 34,581 / 6,491 / 17,778  \\
OrganCMNIST & Abdominal CT & MC (11) & 23,660 & 13,000 / 2,392 / 8,268  \\
OrganSMNIST & Abdominal CT & MC (11) & 25,221 & 13,940 / 2,452 / 8,829 \\
\bottomrule
\end{tabular}
}
\end{table*}

\section{Experiments}
\subsection{Datasets}
We utilized 17 publicly available medical image datasets (detailed in Table \ref{tab:dataset}), all of which are multi-center, to comprehensively evaluate the effectiveness and potential of MedViTV2 in medical image classification. Additionally, to demonstrate the robustness of our proposed model against simulated artifacts and distribution shifts, we evaluated its performance on MedMNIST-C, an open-source benchmark dataset derived from the MedMNIST collection, which encompasses 12 datasets and 9 imaging modalities.

\textbf{MedMNIST}~\cite{yang2023medmnist} repository comprises 12 pre-processed datasets containing OCT, X-ray, CT, and ultrasound images. These datasets support various classification tasks, including  ,multi-class, ordinal, multi-label, binary classification, and regression. Their sizes range from a minimum of approximately 100 images to over 100,000. As depicted in Table~\ref{tab:dataset}, the breadth and variety of the datasets make them particularly conducive to classification research. Pre-processing and partitioning into training, validation, and test subsets follow the procedures outlined in~\cite{yang2021medmnist}.

\textbf{Fetal-Planes-DB}~\cite{burgos2020evaluation}. This dataset is a comprehensive, publicly available collection of maternal-fetal ultrasound images, containing over 12,400 grayscale images from 1,792 patients. Gathered in real clinical settings at BCNatal, Barcelona, it is categorized into six groups: common fetal anatomical planes (Brain, Thorax, Femur, and Abdomen), the maternal cervix, and a general "Other" category. Brain images are further subdivided into three detailed subcategories (Trans-ventricular, Trans-cerebellar, and Trans-thalamic) for fine-grained analysis. Each image was meticulously labeled by expert clinicians and anonymized to ensure patient confidentiality.

\textbf{CPN X-ray}~\cite{kumar2022litecovidnet, shastri2022cheximagenet} This dataset is a structured medical image repository designed to support research and clinical advancements in detecting and classifying COVID-19 and pneumonia using deep learning. Organized into three subfolders including PNEUMONIA, COVID, and NORMAL, it contains a total of 5,228 preprocessed and resized grayscale images in PNG format, each with dimensions of $256\times256$ pixels. The dataset includes 1,626 images of COVID-19 cases, 1,802 normal cases, and 1,800 pneumonia cases.

\textbf{Kvasir}~\cite{pogorelov2017kvasir}. This dataset is a publicly available collection of 4,000 annotated images designed to advance research in the automatic detection and classification of gastrointestinal diseases. Curated by medical experts, it contains eight distinct classes, including anatomical landmarks (e.g., Z-line, cecum, pylorus), pathological findings (e.g., esophagitis, polyps, ulcerative colitis), and endoscopic procedures related to polyp removal. The images, captured during real endoscopic examinations at Vestre Viken Health Trust in Norway, vary in resolution from $720\times576$ to $1920\times1072$ pixels and are organized into separate class-specific folders.

\textbf{PAD-UFES-20}~\cite{pacheco2020pad}. This dataset is a comprehensive collection designed to assist in skin cancer detection, particularly in remote or under-resourced areas. It consists of 2298 clinical images of skin lesions from 1373 patients, collected via smartphones, alongside up to 21 clinical data features for each patient. The dataset includes six diagnostic categories, three skin cancers (SCC, BCC, MEL), and three skin diseases, with 58.4\% of lesions biopsy-proven, including all skin cancer cases. Key attributes in the metadata include patient demographics, lesion characteristics (e.g., itchiness, diameter, elevation), and historical health data (e.g., family cancer history).

\begin{figure*}[t]
    \centering
    \includegraphics[width=\textwidth]{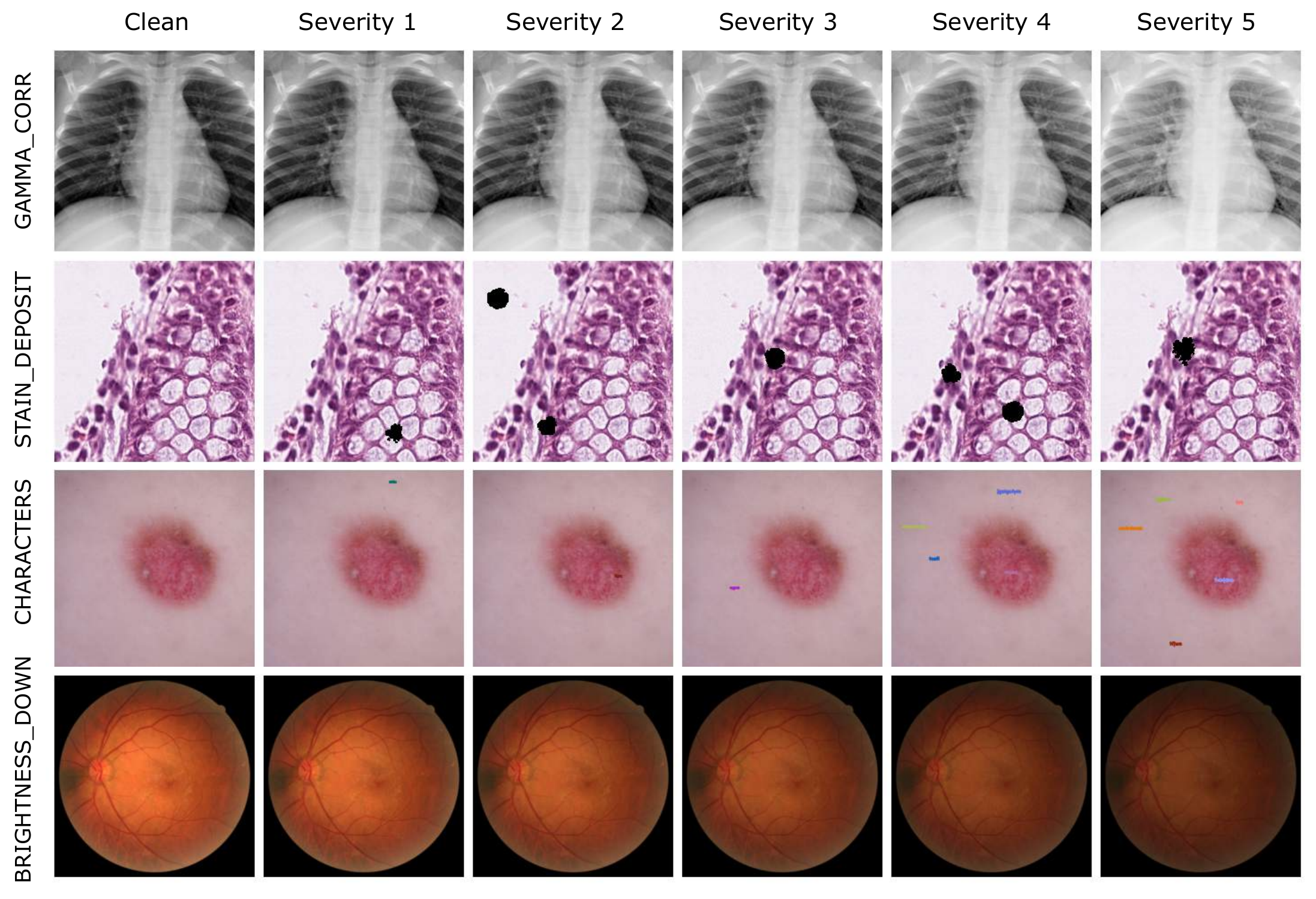}
    \caption{{Overview of the MedMNIST-C Benchmark.} Examples of four artifacts simulated on PneumoniaMNIST, PathMNIST, DermaMNIST, and RetinaMNIST (from top to bottom). Each artifact is applied at five increasing levels of severity.}
    \label{fig:medmnistc}
\end{figure*}

\textbf{MedMNIST-C}~\cite{di2024medmnist}. This is a comprehensive benchmark dataset designed to evaluate the robustness of deep learning algorithms in medical image analysis. It extends just test set of the MedMNIST+ collection by incorporating task-specific and modality-aware image corruptions, simulating real-world artifacts and distribution shifts commonly encountered in medical imaging. Covering 12 datasets and 9 imaging modalities, MedMNIST-C provides a structured framework for testing model performance under diverse conditions, including noise, blur, color alterations, and task-specific distortions.

\begin{figure*}[ht]
    \centering
    \includegraphics[width=\textwidth]{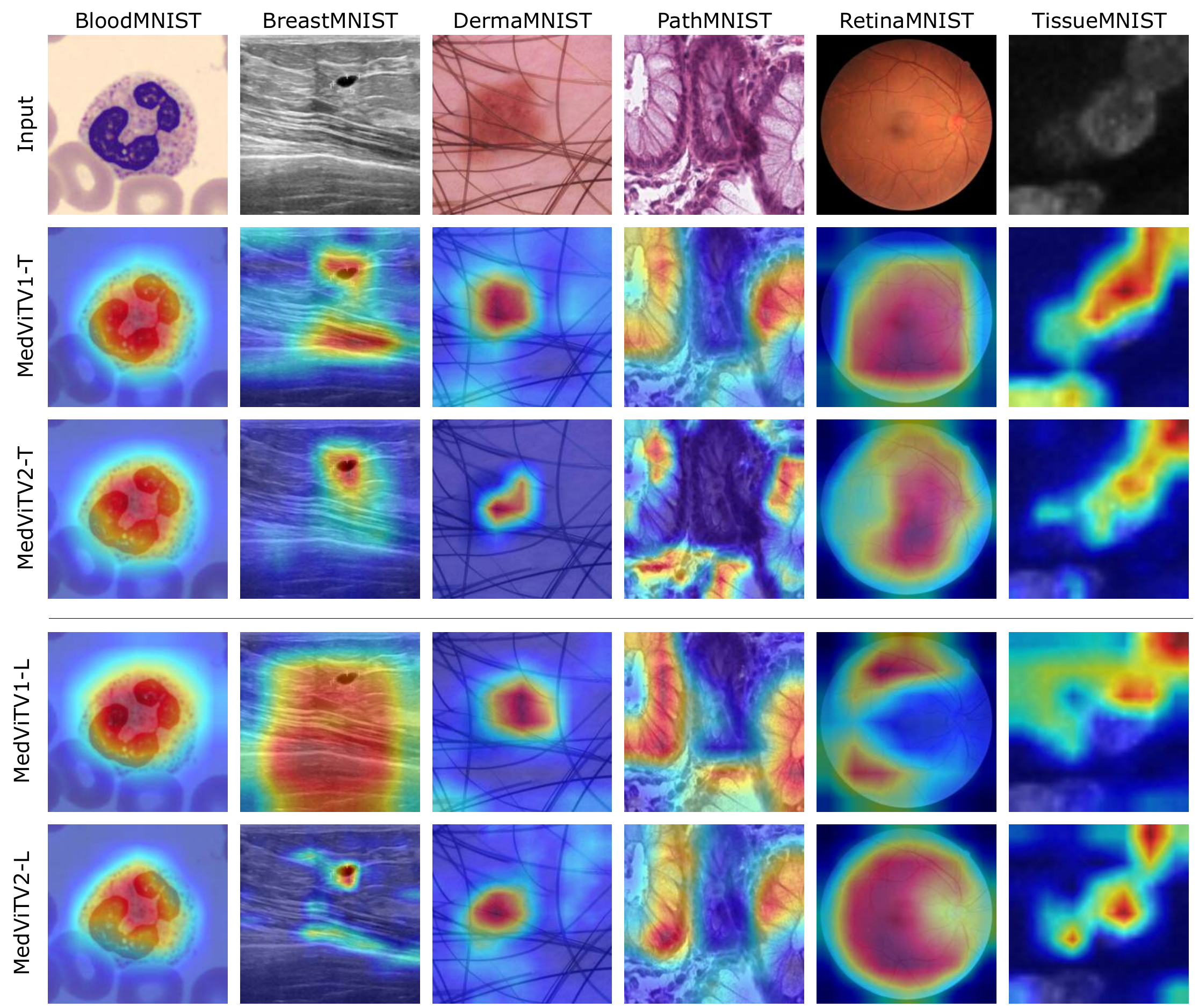}
    \caption{{Grad-Cam heatmap visualization}. We present heatmaps generated from the last three layers of MedViTV1-T, MedViTV2-T, MedViTV1-L, and MedViTV2-L, respectively. Specifically, we utilize the final GFP, LGP, and normalization layers in these models to produce the heatmaps using Grad-CAM.}
    \label{fig:heatmap}
\end{figure*}

\begin{table*}[!htb]
\renewcommand\arraystretch{1.25}
    \caption{ The comparison results of the proposed method on MedMNIST2D are presented in terms of AUC and ACC, with the best results highlighted in bold. Metrics marked with a dash were not reported in their study.}
    \vspace{-15pt}
    \label{tab:2DResults}
    \begin{center}
    \resizebox{1\textwidth}{!}{%
        \begin{tabular}{l||cc||cc||cc||cc||cc||cc}
            \toprule
            \multirow{2.5}{*}{\textbf{Methods}} &
            \multicolumn{2}{c||}{\textbf{PathMNIST}} &
            \multicolumn{2}{c||}{\textbf{ChestMNIST}} &
            \multicolumn{2}{c||}{\textbf{DermaMNIST}} &
            \multicolumn{2}{c||}{\textbf{OCTMNIST}} &
            \multicolumn{2}{c||}{\textbf{PneumoniaMNIST}} &
            \multicolumn{2}{c}{\textbf{RetinaMNIST}} \\ \cline{2-13}
            & AUC & ACC & AUC & ACC & AUC & ACC & AUC & ACC & AUC & ACC & AUC & ACC\\ \midrule
            ResNet-18 (28)~\cite{ResNet}     & 0.983 & 0.907 & 0.768 & 0.947 & 0.917 & 0.735 & 0.943 & 0.743 & 0.944 & 0.854 & 0.717 & 0.524 \\
            ResNet-18 (224)~\cite{ResNet}        & 0.989 & 0.909 & 0.773 & 0.947 &  {0.920} & 0.754 & 0.958 & 0.763 & 0.956 & 0.864 & 0.710 & 0.493 \\
            ResNet-50 (28)~\cite{ResNet}         & 0.990 & 0.911 & 0.769 & 0.947 & 0.913 & 0.735 & 0.952 & 0.762 & 0.948 & 0.854 & 0.726 & 0.528  \\
            ResNet-50 (224)~\cite{ResNet}        & 0.989 & 0.892 & 0.773 &  0.948 & 0.912 & 0.731 & 0.958 & {0.776} & 0.962 & 0.884 & 0.716 & 0.511 \\
            auto-sklearn~\cite{feurer2015efficient}         & 0.934 & 0.716 & 0.649 & 0.779 & 0.902 & 0.719 & 0.887 & 0.601 & 0.942 & 0.855 & 0.690 & 0.515  \\
            AutoKeras~\cite{jin2019auto}           & 0.959 & 0.834 & 0.742 & 0.937 & 0.915 & 0.749 & 0.955 & 0.763 & 0.947 & 0.878 & 0.719 & 0.503 \\
            Google AutoML~\cite{bisong2019google}  & 0.944 & 0.728 &  0.778 &  0.948 & 0.914 &  0.768 & {0.963} & 0.771 &  0.991 &  0.946 &  0.750 &  0.531 \\
            MedViTV1-T~\cite{manzari2023medvit}  &  {0.994} &  {0.938} &  0.786 &  {0.956} & 0.914 &  0.768 &  {0.961} & 0.767 &   {0.993} &   {0.949} &  0.752 &  0.534 \\
            MedMamba-T~\cite{yue2024medmamba} &  {0.997} &  {0.953} & - & - & 0.917 & 0.779 & 0.992 & 0.918 & 0.965 & 0.899 & 0.747 & 0.543 \\
            \rowcolor{gray!20}
            \textbf{MedViTV2-T} &  \textbf{0.998} &  \textbf{0.959} & \textbf{0.791} & \textbf{0.963} & \textbf{0.931} & \textbf{0.781} &  \textbf{0.993} & \textbf{0.927} & \textbf{0.995} & \textbf{0.951} & \textbf{0.761} & \textbf{0.547} \\
            \midrule

            MedViTV1-S~\cite{manzari2023medvit}  &  {0.993} &  {0.942} &   {0.791} &  0.954 &  {0.937} &  {0.780} &  0.960 &  {0.782} &   {0.995} &   {0.961} &   {0.773} &   {0.561} \\
            MedMamba-S~\cite{yue2024medmamba} & 0.997 & 0.955 & - & - & 0.924 & 0.758 & 0.991 & 0.929 & 0.976 & 0.936 & 0.718 & 0.545 \\
            \rowcolor{gray!20}
            \textbf{MedViTV2-S} &  \textbf{0.998} &  \textbf{0.965} &  \textbf{0.803} & \textbf{0.964} & \textbf{0.946} & \textbf{0.792} &  \textbf{0.994} & \textbf{0.942} & \textbf{0.996} & \textbf{0.965} & \textbf{0.780} & \textbf{0.562} \\
            \midrule

            MedMamba-B~\cite{yue2024medmamba} & 0.999 & 0.964 & - & -& 0.925 & 0.757 & 0.996 & 0.927 & 0.973 & 0.925 & 0.715 & 0.553 \\
            \rowcolor{gray!20}
            \textbf{MedViTV2-B} &  \textbf{0.999} &  \textbf{0.970} & \textbf{0.815} & \textbf{0.964} &
            \textbf{0.948} & \textbf{0.808} &  \textbf{0.996} & \textbf{0.944} & \textbf{0.996} & \textbf{0.969} & \textbf{0.783} & \textbf{0.575}\\
            \midrule

            MedViTV1-L~\cite{manzari2023medvit}  & 0.984 & 0.933 &   {0.805} &   {0.959} & {0.920} &  {0.773} &  0.945 & 0.761 &  0.991 &  0.921 &   {0.754} &   {0.552} \\
            %MedMamba-X & 0.999 & 0.962 & - & - & 0.918 & 0.751 & 0.993 & 0.928 & 0.964 & 0.910 & 0.719 & 0.57 \\
            \rowcolor{gray!20}
            \textbf{MedViTV2-L} &   \textbf{0.999} &  \textbf{0.977} & \textbf{0.823} & \textbf{0.967} & \textbf{0.950} & \textbf{0.817} &  \textbf{0.996} & \textbf{0.952} & \textbf{0.997} & \textbf{0.973} & \textbf{0.785} & \textbf{0.578}\\
            \midrule \midrule
            \multirow{2.5}{*}{\textbf{Methods}} &
            \multicolumn{2}{c||}{\textbf{BreastMNIST}} &
            \multicolumn{2}{c||}{\textbf{BloodMNIST}} &
            \multicolumn{2}{c||}{\textbf{TissueMNIST}} &
            \multicolumn{2}{c||}{\textbf{OrganAMNIST}} &
            \multicolumn{2}{c||}{\textbf{OrganCMNIST}} &
            \multicolumn{2}{c}{\textbf{OrganSMNIST}} \\ \cline{2-13}
             & AUC & ACC & AUC & ACC & AUC & ACC & AUC & ACC & AUC & ACC & AUC & ACC \\ \midrule
            ResNet-18 (28)~\cite{ResNet}          & 0.901 & 0.863 &  {0.998} & 0.958 & 0.930 & 0.676 & {0.997} & 0.935 & 0.992 & 0.900 & 0.972 & 0.782 \\
            ResNet-18 (224)~\cite{ResNet}         & 0.891 & 0.833 &  {0.998} &  {0.963} & 0.933 & 0.681 & { 0.998} &  { 0.951} &  {0.994} &  { 0.920} & 0.974 & 0.778 \\
            ResNet-50 (28)~\cite{ResNet}          & 0.894 & 0.838 &  {0.998} &  {0.963} & 0.928 & 0.672 & 0.997 & 0.938 & 0.992 & 0.907 & 0.974 & 0.787 \\
            ResNet-50 (224)~\cite{ResNet}         & 0.889 & 0.828 &  {0.998} &  {0.963} & 0.933 & 0.683 & 0.997 & 0.940 & 0.993 & 0.913 & 0.975 & 0.782 \\
            auto-sklearn~\cite{feurer2015efficient}        & 0.707 & 0.715 & 0.973 & 0.907 & 0.926 & 0.653 & 0.983 & 0.896 & 0.970 & 0.873 & 0.937 & 0.748 \\
            AutoKeras~\cite{jin2019auto}           & 0.841 & 0.790 &  {0.998} & 0.962 & 0.934 & 0.677 & 0.997 & 0.937 & 0.993 & 0.914 & 0.974 & 0.772 \\
            Google AutoML~\cite{bisong2019google}  & 0.906 & 0.859 &  {0.998} &  {0.965} & 0.933 & 0.675 & 0.997 & 0.937 & 0.992 & 0.904 & 0.970 & 0.769 \\

            MedViTV1-T~\cite{manzari2023medvit}  &  {0.923} &  \textbf{0.897} &  {0.998} &  {0.965} & 0.931 & 0.673 &  {0.998} &  {0.951} & 0.993 & 0.912 & 0.973 & 0.778 \\
            MedMamba-T~\cite{yue2024medmamba} & 0.825 & 0.853 & 0.999 & 0.978 & -     & -     & 0.998 & 0.946 & 0.997 & 0.927 & 0.982 & 0.819\\
            \rowcolor{gray!20}
            \textbf{MedViTV2-T} & \textbf{0.944} & {0.882} & \textbf{0.999} & \textbf{0.979} & \textbf{0.936} & \textbf{0.696} & \textbf{0.998} & \textbf{0.958} & \textbf{0.997} & \textbf{0.935} & \textbf{0.985} & \textbf{0.824} \\
            \midrule

            MedViTV1-S~\cite{manzari2023medvit}  &  {0.925} &  \textbf{0.901} &   {0.998} &  {0.965} &  {0.938} &   {0.686} &  {0.998} &  {0.952} &   {0.994} &   {0.920} &   {0.975} &  0.786 \\
            MedMamba-S~\cite{yue2024medmamba} & 0.806 & 0.853 & 0.999 & 0.984 & - & - & 0.999 & 0.959 & 0.997 & 0.944 & 0.984 & 0.833\\
            \rowcolor{gray!20}
            \textbf{MedViTV2-S} &  \textbf{0.947} &  {0.895} & \textbf{0.999} & \textbf{0.985} & \textbf{0.939} & \textbf{0.705} & \textbf{0.999} & \textbf{0.966} & \textbf{0.998} & \textbf{0.950} & \textbf{0.986} & \textbf{0.839}\\
            \midrule

            MedMamba-B~\cite{yue2024medmamba} & 0.849 & 0.891 & 0.999 & 0.983 & - & - & 0.999 & 0.964 & 0.997 & 0.943 & 0.984 & 0.834 \\
            \rowcolor{gray!20}
            \textbf{MedViTV2-B} &  \textbf{0.949} &  \textbf{0.904} & \textbf{0.999} & \textbf{0.985} & \textbf{0.942} & \textbf{0.711} & \textbf{0.999} & \textbf{0.969} & \textbf{0.998} & \textbf{0.953} & \textbf{0.987} & \textbf{0.844}\\
            \midrule

            MedViTV1-L~\cite{manzari2023medvit}  & 0.918 & 0.885 &  {0.998} & 0.964 &  {0.937} &  {0.683} &  {0.998} & {0.951} &  {0.994} &  {0.920} &  {0.975} &  {0.787} \\
            %MedMamba-X & 0.898 & 0.853 & 0.999 & 0.984 & - & - & 0.999 & 0.967 & 0.998 & 0.947 & 0.983 & 0.830\\
            \rowcolor{gray!20}
            \textbf{MedViTV2-L} & \textbf{0.953} &  \textbf{0.910} & \textbf{0.999} & \textbf{0.987} & \textbf{0.943} & \textbf{0.716} & \textbf{0.999} & \textbf{0.973} & \textbf{0.999} & \textbf{0.961} & \textbf{0.987} & \textbf{0.851}\\

            \bottomrule
        \end{tabular}}
    \end{center}
    \vspace{-15pt}
\end{table*}

\subsection{Implementation Details}
\label{sec:2.5}
Our experiments on medical image classification were conducted using the PAD-UFES-20, Fetal-Planes-DB, CPN X-ray, Kvasir, ISIC2018, and MedMNIST dataset, which comprises 16 standardized datasets derived from comprehensive medical resources, encompassing a wide range of primary data modalities representative of medical images. To ensure fairness and objectivity on MedMNIST datasets, we adhered to the same training configurations as MedMNISTv2~\cite{yang2023medmnist} and MedViTV1~\cite{manzari2023medvit}, without modifying the original settings. Specifically, all MedViT variants were trained for 100 epochs on an NVIDIA A100 GPU with 40 GB of VRAM, using a batch size of 128. The images were resized to 224 x 224 pixels. We used the AdamW optimizer~\cite{AdamW} with an initial learning rate of 0.001, which was decayed by a factor of 0.1 at the 50th and 75th epochs. Additionally, we introduced four different MedViT models: MedViTV2-T, MedViTV2-S, MedViTV2-B, and MedViTV2-L, as shown in Table~\ref{tab:configs}. All models were configured with the optimal settings determined in Section \ref{sec:strategy} and were trained separately for each dataset.

During the training of the NonMNIST datasets (PAD-UFES-20, Fetal-Planes-DB, CPN X-ray, ISIC2018, and Kvasir), we adhered strictly to the training configurations outlined in Medmamba~\cite{yue2024medmamba}. The MedViT variants underwent training for 150 epochs, utilizing a batch size of 64. Images were resized to 224 x 224 pixels. Furthermore, we utilized the AdamW optimizer, setting the initial learning rate at 0.0001, with B1 at 0.9, B2 at 0.999, and a weight decay of 1e-4. Cross-Entropy Loss was employed to optimize the model parameters.

We employed MedViTV2-S for the MedMNIST-C datasets. Since MedMNIST-C represents an expansion of the MedMNISTv2~\cite{yang2023medmnist} test set, each model initially required training on the MedMNIST train set before being evaluated on the robustness benchmark of the MedMNIST-C. As the results presented in Table \ref{tab:corruptions} were borrowed from this study \cite{di2024medmnist}, we adhered to the training procedures outlined therein, which are consistent with those used for MedMNIST.

\subsection{Evaluation Metrics}
Since our study involves three different dataset collections, we utilize the standard evaluation metrics for each as follows:
Firstly, for the MedMNIST collection, we use Accuracy (ACC) and Area Under the ROC Curve (AUC), as reported in the original publications~\cite{yang2023medmnist, yang2021medmnist}.
Secondly, for NonMNIST collection (Fetal-Planes-DB, CPN X-ray, Kvasir, and PAD-UFES-20 datasets), we report Accuracy, Precision, Sensitivity, Specificity, F1-score, and AUC, in line with the standard metrics described in the original publication~\cite{yue2024medmamba}.

Finally, for MedMNIST-C, we report a distinct set of metrics that require further elaboration. Specifically, we use balanced Accuracy (\texttt{bACC}), which is applicable to both binary and multi-class classification tasks. \texttt{bACC} is calculated as the arithmetic mean of sensitivity and specificity and is particularly useful for handling imbalanced datasets.
MedMNIST-C serves as a corrupted version of MedMNIST, so we use the notation $bACC_{\text{clean}}$ to denote the balanced accuracy on the original MedMNIST dataset, while $bACC$ represents the balanced accuracy on MedMNIST-C. Given the diverse imbalance ratios across the MedMNIST-C datasets, we follow the approach of~\cite{di2024medmnist}, using the Balanced Error (\textit{i.e.,} $1 - \texttt{bACC}$).
We first calculate the clean, balanced error ($\texttt{BE}_{\text{clean}}$) using the MedMNIST test set. Then, for each corruption $c \in C_d$ and severity level $s$ (an integer ranging from 1 to 5), we compute the balanced error ($\texttt{BE}_{s,c}$). Here, $C_d$ denotes the set of all corruptions associated with a specific dataset $d$ (\textit{e.g.,} $C_{\text{\tiny{derma}}} = \{\textit{defocus}, \dots, \textit{contrast}+, \textit{contrast}-, \dots, \textit{characters}\}$).
Next, we average the errors across all severity levels and normalize them using AlexNet's errors to derive the corruption-specific balanced error ($\texttt{BE}_c$), as formalized in Equation~\ref{bce_cf}:

\begin{equation}
\texttt{BE}_c = \frac{\sum_{s=1}^5 \texttt{BE}_{s,c}}{\sum_{s=1}^5 \texttt{BE}_{s,c}^{\text{AlexNet}}}
\label{bce_cf}
\end{equation}

To further evaluate robustness, we measure the relative balanced error ($\texttt{rBE}_c$) to quantify the performance drop relative to the clean test set, as shown in Equation~\ref{rbce_cf}:

\begin{equation}
\texttt{rBE}_c = \frac{\sum_{s=1}^5 (\texttt{BE}_{s,c} - \texttt{BE}_{\text{clean}})}{\sum_{s=1}^5 (\texttt{BE}_{s,c}^{\text{AlexNet}} - \texttt{BE}_{\text{clean}}^{\text{AlexNet}})}
\label{rbce_cf}
\end{equation}

Finally, we average $\texttt{rBE}_c$ across all corruptions to compute the overall relative balanced error ($\texttt{rBE}$). This metric is crucial for assessing the robustness of models, as it reflects the degree of performance degradation under distribution shifts, with the goal of minimizing this drop.

\begin{table*}[!t]
	\footnotesize
	\caption{The performance comparison between MedViTV2-T and reference models on PAD-UFES-20 and ISIC2018 datasets. The \textbf{bold} font represents the best performance and \underline{underline} indicates the second-best. Red highlights models specifically designed for medical image classification. }
    \vspace{-15pt}
    \label{tab:PAD}
	\begin{center}
 \resizebox{1\textwidth}{!}{%
		\begin{tabular}{c|l|ccccccc}
			\toprule\hline
\textbf{Dataset} &\textbf{Model} &\textbf{FLOPs} &\textbf{\#Param} & \textbf{Precision(\%)}   & \textbf{Sensitivity(\%)}   & \textbf{Specificity(\%)}   & \textbf{F1(\%)}  & \textbf{OA(\%)}  \\ \hline
   \multirow{11}{*}{\rotatebox{90}{PAD-UFES-20}}

&Swin-T\cite{liu2021swin}               &4.5 G	&27.5 M	&38.2	&41.1	&\underline{90.6}	&39.5	&60.5	\\
&ConvNeXt-T\cite{57}          &4.5 G	&27.8 M	&37.2	&33.6	&88.9	&33.7	&54.3	\\
&Repvgg-a1\cite{100}            &2.6 G	&12.8 M	&34.7	&37.7	&89.8	&35.9	&56.7	\\
&Mobilevitv2-200\cite{101} &5.6 G	&17.4 M	&33.9	&32.9	&88.0	&32.2	&49.9	 \\
&EdgeNext-base\cite{102}
&2.9 G	&17.9 M	&35.0	&36.4	&89.9	&34.6	&57.6	\\

&Nest-tiny\cite{103}
&5.8 G	&16.7 M	&49.9	&45.5	&91.3	&42.3	&\underline{63.5}	\\

&Mobileone-s4\cite{104}
&3.0 G	&12.9 M	&35.9	&32.2	&87.9	&32.3	&49.3	\\

&Cait-xxs36\cite{105}
&3.8 G	&17.1 M	&37.1	&37.8	&90.0	&37.0	&58.6	\\
&VMamba-T\cite{42}
&4.4 G	&22.1 M	&53.2	&40.6	&90.0	&41.6	&59.3	\\
&\textcolor{red}{HiFuse-T}\cite{huo2024hifuse}
&8.1 G	&82.5 M	&\underline{55.3}	& \underline{61.4}	&90.1	&57.5	&61.4	\\
&\textcolor{red}{MedMamba-T}\cite{yue2024medmamba}
&4.5 G &14.5 M	&38.4	&36.9	&89.9	&35.8	&58.8	\\
&\textcolor{red}{MedViTV1-T}\cite{manzari2023medvit}
&11.7 G &31.1 M	&53.3	&59.8	&90.4	&56.2	&59.8	\\
\rowcolor{gray!20}
\cellcolor{white}& \textcolor{red}{MedViTV2-T}    & 5.1 G & 12.3 M	& \textbf{63.6}	& \textbf{62.5}	& \textbf{91.7}	& \textbf{61.2}	& \textbf{63.6}\\

\hline \hline
   \multirow{11}{*}{\rotatebox{90}{ISIC2018}}

&Swin-T\cite{liu2021swin}               &4.5 G	&27.5 M	&60.7	&66.1	&91.5	&61.9	&66.1	\\
&ConvNeXt-T\cite{57}          &4.5 G	&27.8 M	&65.3	&67.1	&91.6	&63.2	&67.1	\\
&Repvgg-a1\cite{100}            &2.6 G	&12.8 M	&69.7	&71.6	&92.5	&68.3	&71.6	\\
&Mobilevitv2-200\cite{101} &5.6 G	&17.4 M	&66.4	&68.1	&92.0	&65.2	&68.1	 \\
&EdgeNext-base\cite{102}
&2.9 G	&17.9 M	&64.3	&67.7	&91.7	&64.5	&67.7	\\

&Nest-tiny\cite{103}
&5.8 G	&16.7 M	&67.6	&69.1	&91.3	&64.2	&69.1	\\

&Mobileone-s4\cite{104}
&3.0 G	&12.9 M	&70.0	&72.2	&93.0	&70.0	&72.2	\\

&Cait-xxs36\cite{105}
&3.8 G	&17.1 M	&56.6	&63.9	&90.1	&58.4	&63.9	\\
&VMamba-T\cite{42}
&4.4 G	&22.1 M	&70.5	&72.5	&92.8	&70.3	&72.5	\\
&\textcolor{red}{HiFuse-T}\cite{huo2024hifuse}
&8.1 G	&82.5 M	&\underline{74.8}	&\underline{75.5}	&\underline{93.7}	&\underline{73.9}	&\underline{75.6}	\\
&\textcolor{red}{MedMamba-T}\cite{yue2024medmamba}
&4.5 G &14.5 M	&72.2	&74.1	&93.4	&72.3	&74.0	\\
&\textcolor{red}{MedViTV1-T}\cite{manzari2023medvit}
&11.7 G &31.1 M	&71.5	&72.4	&92.4	&69.4	&72.4	\\
\rowcolor{gray!20}
\cellcolor{white}& \textcolor{red}{MedViTV2-T}    & 5.1 G & 12.3 M	& \textbf{76.1}	& \textbf{77.1}	& \textbf{94.4}	& \textbf{76.2}	& \textbf{77.1}	\\

\hline\bottomrule
		\end{tabular}}
		\label{tab3}
	\end{center}
\end{table*}

\begin{table*}[!t]
\footnotesize
	\caption{The performance comparison between MedViTV2-S and reference models on CPN and Kvasir datasets. The \textbf{bold} font represents the best performance and \underline{underline} indicates the second-best. Red highlights models specifically designed for medical image classification.}
    \vspace{-15pt}
    \label{tab:CPN-Kvasir}
	\begin{center}
 \resizebox{1\textwidth}{!}{%
		\begin{tabular}{c|l|ccccccc}
			\toprule\hline
\textbf{Dataset} &\textbf{Model} &\textbf{FLOPs} &\textbf{\#Param} & \textbf{Precision(\%)}   & \textbf{Sensitivity(\%)}   & \textbf{Specificity(\%)}   & \textbf{F1(\%)}  & \textbf{OA(\%)}  \\ \hline
   \multirow{12}{*}{\rotatebox{90}{CPN X-ray}}

&{Swin-S} &8.7 G	&48.8 M	&95.4	&95.5	&97.7	&95.4	&95.4	\\
&{ConvNext-S} &8.7 G	&49.4 M	&95.7	&95.7	&97.8	&95.7	&95.6 	\\
&Convformer-s18 &4.0 G	&24.7 M	&95.9	&95.8	&97.8	&95.8	&95.7	 \\
&TNT-s &5.2 G	&23.3 M	&93.4	&93.4	&96.6	&93.4	&93.2	 \\
&Caformer-s18 &4.1 G	&24.3 M	&95.5	&95.5	&97.7	&95.5	&95.4	 \\
&PvtV2-b2 &4.0 G	&24.8 M	&96.3	&96.2	&98.1	&96.2	&96.2	 \\
&Davit-tiny &4.5 G	&27.6 M	&95.1	&95.2	&97.5	&95.1	&95.1	 \\
&Deit-small &4.6 G	&21.7 M	&95.2	&95.1	&97.5	&95.1	&95.1	 \\

&EfficientNetV2-s &8.3 G	&20.2 M	&95.8	&95.7	&97.8	&95.7	&95.7	\\

&Coat-small &12.6 G	&21.4 M	&94.3	&94.2	&97.0	&94.2	&94.1	\\
&{VMamba-S} &9.0 G	&43.7 M	&96.8	&96.8	&98.3	&96.8	&96.8	 \\
&\textcolor{red}{HiFuse-S} &8.8 G	&93.8 M	&95.5	&95.4	&97.7	&95.4	&95.4	 \\
&\textcolor{red}{MedMamba-S}    &6.1 G	&22.8 M	&\underline{97.4}	&\underline{97.4}	&\underline{98.6}	&\underline{97.4}	&\underline{97.3}	\\
&\textcolor{red}{MedViTV1-S}    &16.7 G	&44.4 M	&96.7	&96.8	&98.3	&96.7	&96.7	\\
\rowcolor{gray!20}
\cellcolor{white}&\textcolor{red}{MedViTV2-S}    &  7.6 G	& 29.6 M	&\textbf{98.2}	&\textbf{98.2}	&\textbf{99.1}	&\textbf{98.2}	&\textbf{98.2}	\\
\hline\hline
\multirow{12}{*}{\rotatebox{90}{Kvasir}}
&{Swin-S}
&8.7 G	&48.8 M	&78.4	&78.0	&96.9	&77.3	&78.0	\\
&{ConvNext-S}
&8.7 G	&49.4 M	&75.6	&74.8	&96.1	&74.8	&74.8	 \\
&Convformer-s18
&4.0 G	&24.7 M	&76.4	&75.8	&96.5	&75.6	&75.8	\\
&TNT-s
&5.2 G	&23.3 M	&76.5	&76.2	&96.6	&75.7	&76.2	\\
&Caformer-s18
&4.1 G	&24.3 M	&73.6	&73.7	&96.2	&73.5	&73.7	\\

&PvtV2-b2
&4.0 G	&24.8 M	&75.7	&75.6	&96.5	&75.3	&75.6	\\

&Davit-tiny
&4.5 G	&27.6 M	&73.8	&73.6	&96.2	&73.0	&73.6	\\

&Deit-small
&4.6 G	&21.7 M	&78.2	&78.1	&96.8	&77.9	&78.1	\\

&EfficientNetV2-s
&8.3 G	&20.2 M	&78.7	&78.1	&96.8	&78.1	&78.2	\\

&Coat-small
&12.6 G	&21.4 M	&74.2	&73.5	&96.2	&73.1	&73.5	\\
&{VMamba-S}
&9.0 G	&43.7 M	&77.6	&77.3	&96.8	&77.1	&77.3	\\
&\textcolor{red}{HiFuse-S}
&8.8 G	&93.8 M	&\underline{81.4}	&\underline{81.0}	&\underline{97.3}	&\underline{80.5}	&\underline{81.0}	\\
&\textcolor{red}{MedMamba-S}   &6.1 G	&22.8 M	&79.4	&79.3	&97.0	&79.2	&79.3	\\
&\textcolor{red}{MedViTV1-S}   &16.7 G	&44.4 M	&\underline{81.4}	&80.2	&97.2	&79.6	&80.2	\\
\rowcolor{gray!20}
\cellcolor{white}&\textcolor{red}{MedViTV2-S}   & 7.6 G	& 29.6 M	&\textbf{84.0}	&\textbf{82.8}	&\textbf{97.6}	&\textbf{82.5}	&\textbf{82.8}	\\
\hline\bottomrule
		\end{tabular}}
		\label{tab4}
	\end{center}
\end{table*}

\begin{table*}[!t]
\footnotesize
	\caption{The performance comparison between MedViTV2-B and reference models on Fetal-Planes-DB datasets. The \textbf{bold} font represents the best performance and \underline{underline} indicates the second-best. Red highlights models specifically designed for medical image classification.}
    \vspace{-15pt}
    \label{tab:Fetal}
	\begin{center}
 \resizebox{1\textwidth}{!}{%
		\begin{tabular}{c|l|ccccccc}
			\toprule\hline
\textbf{Dataset} &\textbf{Model} &\textbf{FLOPs(G)} &\textbf{\#Param} & \textbf{Precision(\%)}   & \textbf{Sensitivity(\%)}   & \textbf{Specificity(\%)}   & \textbf{F1(\%)}  & \textbf{OA(\%)}  \\ \hline
   \multirow{13}{*}{\rotatebox{90}{Fetal-Planes-DB}}

&{Swin-B}\cite{liu2021swin}
&15.4 G	&86.7 M	&86.1	&84.9	&97.7	&85.4	&89.2	\\
&{ConvNext-B}\cite{57}
&15.4 G	&87.6 M	&85.9	&85.2	&97.7	&85.5	&89.1	\\
&Davit-small\cite{110}
&8.8 G	&48.9 M	&85.9	&84.8	&97.6	&85.3	&88.9	\\
&Mvitv2-base\cite{114}
&10.0 G	&50.7 M	&89.9	&90.1	&98.3	&89.9	&91.9	\\
&EfficientNet-b6\cite{115}
&19.0 G	&40.8 M	&91.2	&91.2	&98.4	&91.1	&92.8	\\
&EfficientNetV2-b\cite{112}
&24.5 G	&52.9 M	&87.6	&89.1	&97.9	&88.3	&90.2	\\
&FocalNet-s\cite{116}
&8.7 G	&49.1 M	&91.7	&90.9	&98.5	&91.2	&92.9	\\
&Twins-SVT-base\cite{chu2021twins}
&8.8 G	&48.9 M	&87.5	&88.4	&97.9	&88.0	&90.3	\\

&Poolformer-m36\cite{118}
&8.8 G	&55.4 M	&82.7	&82.3	&87.4	&82.9	&87.7	\\

&Xcit-s\cite{119}
&9.1 G	&47.3 M	&85.2	&86.1	&97.7	&85.5	&89.1	\\
&GcVit-s\cite{120}
&8.4 G	&50.3 M	&84.5	&84.3	&97.5	&84.3	&88.4	\\

&{VMamba-B}\cite{42}
&15.1 G	&75.2 M	&92.2	&93.4	&98.7	&92.7	&93.8	\\
&\textcolor{red}{HiFuse-B}\cite{huo2024hifuse}
&10.9 G	&127.8 M	&91.9	&91.7	&98.2	&91.8	&91.7	\\
&\textcolor{red}{MedMamba-B}\cite{yue2024medmamba}
&13.4 G	&47.1 M	&92.8	&\underline{93.8}	&\underline{98.8}	&\underline{93.3}	&\underline{94.4}	\\
&\textcolor{red}{MedViTV1-L}\cite{manzari2023medvit}
&21.6 G	&57.6 M	&\underline{93.2}	&93.2	&98.5	&93.2	&93.2	\\
\rowcolor{gray!20}
\cellcolor{white}&\textcolor{red}{MedViTV2-B}    &15.6 G	& 72.3 M	&\textbf{95.6}	&\textbf{95.3}	&\textbf{99.0}	&\textbf{95.3}	&\textbf{95.3}	\\

\hline\bottomrule
		\end{tabular}}
		\label{tab5}
	\end{center}
\end{table*}

\subsection{Performance on MedMNIST}

Table~\ref{tab:2DResults} reports the performance comparison of MedViTV2 with previous SOTA methods in terms of AUC and ACC on each dataset of MedMNIST2D. Compared with the well-known ResNet and MedMamba, the four variants of MedViTV2 (tiny, small, base, and large) significantly improve the ACC and AUC on each dataset. For instance, in the OCTMNIST dataset, MedViTV2-small achieves AUC and ACC improvements of 3.6\% and 16.6\%, respectively, over ResNet-50. Similarly, in the PneumoniaMNIST dataset, MedViTV2-tiny achieves an improvement of 3.0\% in AUC and 5.2\% in ACC over MedMamba-T.
Overall, MedViTV2 demonstrates exceptional performance on medical image classification tasks in the MedMNIST2D benchmark. Significant improvements are observed in all MedMNIST datasets. To illustrate the potential of MedViTV2 more intuitively, Figure~\ref{fig:lineplot} presents the average ACC and FLOPs for all model sizes. Results show that the MedViTV2 variants achieve average ACC values of 86.6\%, 87.6\%, 88.2\%, and 88.8\% for tiny, small, base, and large, respectively. Notably, MedViTV2 addresses our concerns with MedViTV1, which experienced a drop in accuracy when scaled. Additionally, it strikes an optimal balance between accuracy and complexity, making it advantageous for practical deployment in real-world medical applications.

\begin{table*}[!t]
\renewcommand\arraystretch{1.25}
%\footnotesize
	\caption{The performance of MedViTV2-S and the reference models on the MedMNIST-C benchmark is presented. The bACC, rBE, and BE scores (\%) are averaged across all 12 datasets in the MedMNIST-C benchmark. BE scores are reported separately for each corruption category: Digital, Noise, Blur, Color, and Task-Specific (TS). The best results are highlighted in bold.}
    \vspace{-15pt}
	\begin{center}
 \resizebox{0.9\textwidth}{!}{%
    \begin{tabular}{l||ccccc||ccccc}
    \toprule

\multirow{2}{*}{\textbf{Methods}}& \multirow{2}{*}{\textbf{\#Param}}& \multirow{2}{*}{\textbf{bACC$_{\text{clean}}\uparrow$}} &
\multirow{2}{*}{\textbf{bACC $\uparrow$}}&
\multirow{2}{*}{\textbf{rBE $\downarrow$}}&
\multirow{2}{*}{\textbf{BE $\downarrow$}}&
\multicolumn{5}{c}{\textbf{BE} $\downarrow$} \\

\cline{7-11}

 & & &  & &
					  & \textbf{Digital}
					  & \textbf{Noise}
					  & \textbf{Blur}
					  & \textbf{Color}
					  & \textbf{TS}
                      \\ \midrule
AlexNet \cite{Alexnet} & 62.3 M
			 &  78.7 &  62.9 &  100.0 &  100.0 & 100 & 100 & 100 & 100 & 100
\\
R.Net50 \cite{ResNet} & 25.6 M
			 &  75.4 &  56.2 &  166.1 &  131.5 & 177 & 110 & 123 & 148 & 95
\\
D.Net121 \cite{DenseNet}& 8.0 M
			 &  79.8 &  59.4 &  148.4 &  114.8 & 145 & 124 & 100 & 124 & 78
\\
VGG16 \cite{VGG}& 138.4 M
			 &  80.5 &  65.9 &  114.0 &  93.0  & 128 & 87 & 91 & 84 & 80
\\
ViT-B \cite{ViT}& 86.6 M
			 &  78.9 &  72.0 &  \textbf{59.9} &  76.3   & 74 & 50 & 77 & \textbf{80} & 71
\\
\rowcolor{gray!20}
\textbf{MedViTV2-S} & 32.3 M
			 &  \textbf{84.1} &  \textbf{75.2} &  {89.2} &  \textbf{71.1}
			 & \textbf{50} & \textbf{50} & \textbf{57} & {101} & \textbf{64}
\\
\bottomrule
    \end{tabular}}
		\label{tab:corruptions}
	\end{center}
\end{table*}

\begin{table*}[!t]
\renewcommand\arraystretch{1.25}
%\footnotesize
	\caption{\textbf{Ablation experiments on the impact of KAN and DiNA blocks on corrupted TissueMNIST dataset.} The best results are in bold, and the second-best results are underlined.}
    \vspace{-15pt}
	\begin{center}
 \resizebox{0.8\textwidth}{!}{%
		\begin{tabular}{c|l|ccccc|cccc}
		\toprule\hline
                \multirow{2}{*}{\textbf{Size}} &
                \multirow{2}{*}{\textbf{Model}} &
			\multicolumn{2}{c}{\textbf{LFP}} &
            \multicolumn{3}{c|}{\textbf{GFP}} &
			\multicolumn{4}{c}{\textbf{Evaluation metrics}}\\
  \cmidrule(lr){3-4} \cmidrule(lr){5-7}
  &&\textbf{MHCA} &\textbf{Dilated} &\textbf{MLP} &\textbf{LFFN} &\textbf{KAN} &\textbf{FLOPs(G)$\downarrow$} &\textbf{Paras(M)$\downarrow$} &\textbf{bACC$_{\text{clean}}(\%)\uparrow$} &\textbf{bACC(\%) $\uparrow$}

\\ \hline
\multirow{6}{*}{\rotatebox{90}{Tiny}}
&A&\checkmarks &\crossmark	&\checkmarks &\crossmark &\crossmark &5.61	&13.73	&58.2	&44.2\\
&B (MedViTV1-T)&\checkmarks &\crossmark	&\crossmark &\checkmarks &\crossmark &5.82	&11.81	&68.6	&50.1\\
&C&\checkmarks &\crossmark	&\crossmark &\crossmark &\checkmarks &5.48	&12.78	&\underline{71.2}	&\underline{52.5}\\
&D&\crossmark &\checkmarks	&\checkmarks &\crossmark &\crossmark &5.63	&13.26	&56.3	&41.8 \\
&E&\crossmark &\checkmarks	&\crossmark &\checkmarks &\crossmark &5.63	&11.34	&63.1	&48.5 \\
\rowcolor[HTML]{C8FFFD}
\cellcolor{white}&F (MedViTV2-T)&\crossmark &\checkmarks	&\crossmark &\crossmark &\checkmarks &5.50	&12.31	&\textbf{72.7}	&\textbf{56.9} \\ \hline
\multirow{6}{*}{\rotatebox{90}{Large}}
&G&\checkmarks &\crossmark	&\checkmarks &\crossmark &\crossmark &25.23	&178.54	&57.1	&42.5\\
&H (MedViTV1-L)&\checkmarks &\crossmark	&\crossmark &\checkmarks &\crossmark &25.25	&153.76	&67.9	&\underline{53.3}\\
&I&\checkmarks &\crossmark	&\crossmark &\crossmark &\checkmarks &23.82	&179.54	&\underline{72.4}	&50.6\\
&J&\crossmark &\checkmarks	&\checkmarks &\crossmark &\crossmark &26.52	&175.76	&55.1	&39.3 \\
&K&\crossmark &\checkmarks	&\crossmark &\checkmarks &\crossmark &26.54	&142.97	&64.4	&50.8 \\
\rowcolor[HTML]{C8FFFD}
\cellcolor{white}&L (MedViTV2-L)&\crossmark &\checkmarks	&\crossmark &\crossmark &\checkmarks &25.12	&162.90	&\textbf{74.1}	&\textbf{59.1} \\
\hline\bottomrule
		\end{tabular}}
		\label{tab:ablation}
	\end{center}
\end{table*}

\subsection{Performance on NonMNIST}
In this section, we evaluate the performance of the MedViTV2 model against the latest SOTA models, including CNNs, ViTs, Mambas, and hybrid networks, with parameter sizes comparable to our model. The model sizes and reported metrics are based on the original work by Yue et al. \cite{yue2024medmamba}.

The performance comparison in Table \ref{tab:PAD} highlights the superior performance of MedViTV2-tiny compared to several SOTA models across the PAD-UFES-20 dataset. MedViTV2-tiny achieves the highest scores across nearly all evaluation metrics, with a precision of 63.6\%, sensitivity of 62.5\%, specificity of 91.7\%, F1-score of 61.2\%, overall accuracy of 63.6\%, and an AUC of 87.7\%. Notably, this performance is achieved with the lowest computational complexity and parameter size, demonstrating MedViTV2-tiny's remarkable efficiency and effectiveness. This underscores MedViTV2-tiny's capability to provide a lightweight yet highly performant solution, establishing it as a SOTA model for PAD-UFES-20.

Table~\ref{tab:CPN-Kvasir} shows the performance of MedViTV2-small and SOTA models on Kvasir and CPN X-ray. On CPN X-ray, MedViTV2-small achieves the best performance metrics among all models. Compared with recent models, the OA of MedViTV2-small increases by 0.9\% (MedMamba-S), 1.4\% (VMamba-S), 2.8\% (Swin-S), and 2.6\% (ConvNext-S), respectively, while maintaining a competitive parameter size. Similarly, the performance of MedViTV2-small is remarkable on Kvasir. In terms of OA, MedViTV2-small outperforms all reference models by achieving a significant improvement of 3.5\% (MedMamba-S), 5.5\% (VMamba-S), and 8.0\% (ConvNext-S). Regarding AUC, MedViTV2-small achieves the highest result, outperforming all competitors, including the previously top-ranked Deit-small by 0.7\%.

Table~\ref{tab:Fetal} reports the performance of MedViTV2-base and reference models on the Fetal-Planes-DB dataset. MedViTV2-base achieves the best OA and AUC among all models while maintaining the lowest FLOPs. Specifically, compared with counterpart models, MedViTV2-base improves the OA by 0.9\% over MedMamba-B, 1.5\% over VMamba-B, 6.1\% over Swin-B, and 6.2\% over ConvNext-B. Similarly, in terms of AUC, MedViTV2-base achieves an improvement of 0.4\% over MedMamba-B, 0.3\% over VMamba-B, and 1.5\% over both Swin-B and ConvNext-B. These results highlight the substantial advancements MedViTV2-base offers in medical image analysis with superior performance and efficiency.

\subsection{Robustness Evaluation}
As shown in Table \ref{tab:corruptions}, our experiment investigates the robustness performance of MedViTV2-small against image artifacts compared to widely used models in the MedMNIST-C benchmark. The results demonstrate that our model achieves the best robustness while containing only 32.3 million parameters. As anticipated, ViT-B ranks second in robustness, while VGG is the runner-up in clean performance, albeit with the highest number of parameters.
The results also highlight that the degree of robustness varies across different types of corruption. For instance, digital and noise corruptions have the least impact on our model, whereas color corruptions result in a larger performance gap between clean and robust metrics.
An important finding from our study is that digital corruptions have the most significant impact on CNN performance, while color corruptions have more severe effects on ViT models. As a direction for future work, we aim to design a model with robust architectural blocks to address these weaknesses effectively. For more details on the results for each dataset, please refer to this repository\footnote{\url{https://github.com/Omid-Nejati/MedViTV2/tree/main/checkpoints}}.

\subsection{Heatmap Visualization}
\begin{figure*}[t]
\centering
\includegraphics[width=0.99\linewidth]{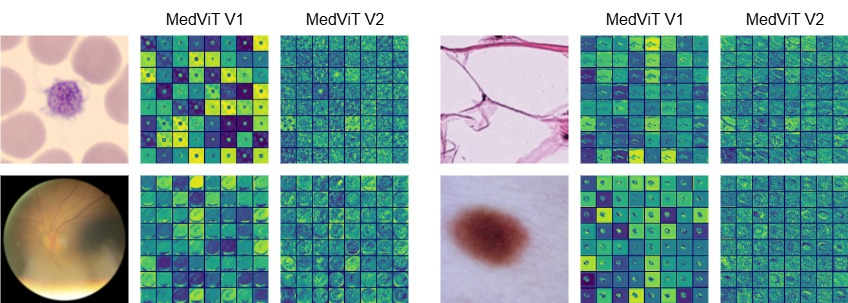}
\caption{
We visualize the activation map for each feature channel, presenting them in small square arrays. For enhanced clarity, each visualization displays 64 distinct channels. Our analysis revealed that the MedViTV1 model is susceptible to a feature collapse issue, evidenced by the presence of inactive or saturated neurons across its channels. To mitigate this problem, we introduce a novel DiNA method designed to enhance feature diversity and expand the receptive field during the training process. This technique is applied to high-dimensional features within every block, which ultimately led to the development of the MedViTV2 architecture.
}
\label{fig:collapse}
\end{figure*}

To gain deeper insight into the learning behavior, we perform a qualitative analysis of the feature space, as shown in Figure \ref{fig:heatmap}. Using both MedViTV1 and MedViTV2 in their tiny and large configurations, we generate heatmaps for several datasets from the MedMNIST benchmark using Grad-CAM \cite{selvaraju2020grad}.
Also, we perform qualitative analysis in the feature space. We visualize the activations in Figure
\ref{fig:collapse}.
An intriguing phenomenon, referred to as ``feature collapse" \cite{woo2023convnext}, is observed in MedViTV1-L for certain datasets, including BreastMNIST, RetinaMNIST, and TissueMNIST. This occurs when many feature maps become saturated or inactive, primarily in the dimension expansion layers of the MedViTV1 block. To address this issue, we propose combining DiNA block to diversify the feature representations during scaling, effectively mitigating feature collapse.
As a result, MedViTV2-L demonstrates improved attention quality, focusing on more relevant areas of the images compared to its smaller version. Notably, MedViTV2-T has only $\sim$12 million parameters and captures critical features, highlighting the most important regions. In contrast, MedViTV2-L, with over ten times the number of parameters, is capable of capturing richer features, resembling segmentation masks in cases such as RetinaMNIST and BreastMNIST.

\section{Ablation Study}
\subsection{Impact of Components}
Can Transformers fuse robust features for medical image classification? To address this question, Table \ref{tab:ablation} examines the effectiveness of different combinations of our components in MedViTV1 and MedViTV2 on corrupted TissueMNIST dataset. We consider both clean and robust accuracy as metrics to identify the best feature extractors for the architecture of MedViTV2. Additionally, we evaluate these components across both tiny and large model sizes to overcome a major limitation of MedViTV1, which struggled to improve accuracy with model scaling.
Rows (B and H) correspond to MedViTV1, which, as noted, suffers a drop in clean bACC at a larger model size. Notably, the inclusion of KAN in various combinations consistently enhances clean accuracy (rows C, F, I, and L). Meanwhile, LFFN demonstrates strong capability in generating robust features, achieving the second-best bACC in a larger model (row H), although it does not improve clean accuracy as effectively as KAN. Finally, the components of MedViTV2, specifically Dilated and KAN (rows F and L), achieve the best performance across both clean and robust accuracy metrics.
So, the answer to the title question is yes: Transformers, when combined with CNNs and efficient blocks such as KAN and LFFN, can effectively fuse robust features for generalized medical image classification.

\begin{table*}[h]
\renewcommand\arraystretch{1.25}
    \caption{\textbf{Memory usage, FLOPs, and receptive field sizes in different attention methods.}
    The receptive field of Neighborhood Attention (NA) grows linearly as the model's depth increases.
    Swin also shows a linear progression in its receptive field expansion.
    Notably, DiNA offers the distinct advantage of \textit{exponentially} expanding its receptive field.
    While self-attention attains the maximal receptive field size, this is accompanied by a quadratic increase in computational operations.
    The upper limit for all reported receptive field sizes is constrained by $n$.}
    \vspace{-15pt}
    \begin{center}
    \resizebox{0.8\textwidth}{!}{%
    \begin{tabular}{l|cc|l|cc}
        \toprule
        \textbf{Model}        & \textbf{Memory usage} & \textbf{FLOPs}        & \textbf{Receptive Field} & TissueMNIST(\%) & TissueMNIST-C(\%)\\
        \hline
        \multirow{2}{*}{\textbf{ViT}}           & $3d^2 + n^2$          & $3nd^2 + 2 n^2 d$     & \multirow{2}{*}{$n$} & \multirow{2}{*}{64.7} & \multirow{2}{*}{46.3}\\
        & (6.2 GB) & (17.6 G) & \\
        \hline
        \multirow{2}{*}{\textbf{Swin}}        & $3d^2 + n k$          & $3nd^2 + 2 n d k$     & \multirow{2}{*}{$\ell k$} & \multirow{2}{*}{66.3} & \multirow{2}{*}{39.7}\\
        & (6.7 GB)& (15.4 G) & \\
        \hline
         \multirow{2}{*}{\textbf{NA}}           & $3d^2 + n k$          & $3nd^2 + 2 n d k$     & \multirow{2}{*}{$\ell (k - 1) + 1$} & \multirow{2}{*}{69.5} & \multirow{2}{*}{49.1}\\
         & (5.0 GB)& (13.7 G) & \\
         \hline
         \multirow{2}{*}{\textbf{DiNA}} & $3d^2 + n k$          & $3nd^2 + 2 n d k$     & \multirow{2}{*}{$\in [\ell (k-1) + 1, k^\ell]$} & \multirow{2}{*}{71.1} & \multirow{2}{*}{54.5}\\
         & (5.2 GB)& (15.6 G) & \\
        \bottomrule
		\end{tabular}}
		\label{tab:receptive}
	\end{center}
\end{table*}

\subsection{DiNa vs. Attention Methods}
To fully appreciate DiNA's effectiveness, particularly in comparison to other models, a thorough analysis of its receptive field (RF) is essential. Table \ref{tab:receptive} provides a comparative overview of receptive field sizes across various attention patterns, along with their associated computational (FLOPs) and memory costs for TissueMNIST and TissueMNIST-C. We determine the receptive field size by considering the number of layers $\ell$, kernel size $k$, and the number of tokens $n$. Initially, Neighborhood Attention (NA) has a receptive field of $k$, which expands by $k-1$ with each subsequent layer while maintaining the central pixel fixed. Conversely, the Window Self Attention employed in the Swin Transformer~\cite{liu2021swin} intrinsically maintains a constant receptive field because its window partitioning prevents interactions between windows, thereby precluding further RF expansion. This limitation is mitigated by pixel-shifted WSA, which facilitates a receptive field expansion of exactly one window per layer, translating to a $k$ expansion per layer.

In stark contrast to the fixed receptive field growth exhibited by NAT and Swin, DiNA's receptive field demonstrates remarkable adaptability and is contingent on the chosen dilation. Its receptive field can span from a linear progression, similar to NAT's $\ell (k - 1) + 1$ (when all dilation values are set to 1), to an exponential growth of $k^\ell$ (when dilation increases progressively). This inherent flexibility in dilation is a primary driver of DiNA's superior performance. Regardless of the dilation strategy, the first layer of DiNA invariably yields a receptive field of size $k$. When sufficiently large dilation values are utilized, each DiNA layer contributes a receptive field of size $k$ per dimension, culminating in an effective receptive field of $k^2$. This allows for DiNA and NA combinations, when paired with optimal dilation values, to achieve an impressive exponential receptive field expansion up to $k^\ell$. This observation is consistent with established knowledge regarding dilated convolutions, which are known to exhibit exponential receptive field growth when using exponentially increasing dilation factors~\cite{hassani2023neighborhood}.
In our experiments, DiNA demonstrates notable improvements, surpassing Swin and NAT by 4.8\% and 1.6\%, respectively, on TissueMNIST. Furthermore, DiNA significantly excels on the corruption benchmark, outperforming Swin and NAT by 14.8\% and 5.4\%, respectively, on TissueMNIST-C.

\begin{table}[h]
    \centering
    \caption{Dilation impact on performance.}
    %\vspace{-15pt}
    \resizebox{0.7\textwidth}{!}{%
    \begin{tabular}{c|cc|cc}
        \toprule
        \textbf{Dilation per level}      & TissueMNIST(\%) & DermaMNIST(\%) & TissueMNIST-C(\%) & DermaMNIST-C(\%)   \\

        \midrule
        1, 1, 1, 1                    & 68.7    & 78.8 & 45.6  & 52.0    \\
        2, 2, 2, 1                    & 69.8    & 79.2 & 49.4  & 60.9    \\
        4, 4, 2, 1                    & 70.4    & 80.1 & 52.3  & 65.5    \\
        8, 4, 2, 1                    & 71.1    & 80.8 & 54.5  & 69.6    \\
        \bottomrule
    \end{tabular}}
    \label{tab:dilationperformance}
\end{table}

\subsection{Dilation Values}
Our findings in Table~\ref{tab:dilationperformance} illustrate the impact of different dilation values on both standard and robust classification performance. It should be highlighted that the progressively increasing dilation sequence $(8, 4, 2, 1)$ was exclusively utilized for the classification task. This is because, from a theoretical standpoint, the input feature maps must have dimensions at least equal to the product of the kernel size and the dilation factor. Therefore, for MedMNIST images at a \(224 \times 224\) resolution, the sequence $(8, 4, 2, 1)$ represents the upper limit of feasible dilation. Should image resolution be increased, higher dilation values could be employed. We investigated a spectrum of dilation configurations within DiNA layers, ranging from the NAT model's uniform $(1, 1, 1, 1)$ to MedViTV2's varied $(8, 4, 2, 1)$, as summarized in Table~\ref{tab:dilationperformance}. Ultimately, we opted for a ``gradual" dilation scheme, where the dilation value incrementally approaches a predetermined maximum.

\begin{table}[h]
\centering
\caption{Comparison of different KAN blocks and MLP method.}
\resizebox{0.9\textwidth}{!}{%
\label{tab:KAN}
\begin{tabular}{l|cccc|cccc}
\toprule
\textbf{Model} &  \textbf{\#Param.} &\textbf{FLOPs} & \textbf{Memory usage} & \textbf{throughput(ms)} & TissueMNIST(\%)& TissueMNIST-C(\%) \\
\midrule
MLP & 78.2 M & 7.5 G & 3.7 GB & 178 & 65.9 & 43.2 \\
EfficientKAN & 90.1 M & 16.8 G &  17.8 GB & 521 & 69.5 & 45.9  \\
FastKAN  & 84.3 M& 16.5 G &  7.6 GB & 309 & 69.7 & 43.4 &  \\
FC-KAN   & 97.6 M & 18.3 G & 19.5 GB & 576 & 70.8 & 46.7 &   \\
Our & 72.3 M&  15.6 G & 5.2 GB & 293 & 71.1 & 54.5 \\
\bottomrule
\end{tabular}}
\end{table}

\subsection{Impact of KAN Block}
To validate the effectiveness of our proposed KAN component, we substituted the KAN component in MedViTV2 with several established blocks, including the MLP from ViT~\cite{ViT}, EfficientKAN\footnote{\url{https://github.com/Blealtan/efficient-kan}}, FastKAN~\cite{li2024kolmogorov}, and FC-KAN~\cite{ta2024fc}. For a rigorous and fair comparison, we consistently applied LFP and GFP in the construction of all models, adhering to the settings outlined in Section~\ref{sec:2.5}. As demonstrated in Table~\ref{tab:KAN}, our proposed KAN achieves the most favorable balance between resource consumption and accuracy across both benchmarks. This empirically confirms the advantages of our KAN component. For example, our KAN outperforms the recent FC-KAN block~\cite{ta2024fc} by 0.3\% on TissueMNIST and a substantial 7.8\% on TissueMNIST-C, while concurrently achieving a significant reduction in memory usage by approximately 73.33\%.

\begin{figure*}[t]
\centering
\includegraphics[width=0.75\linewidth]{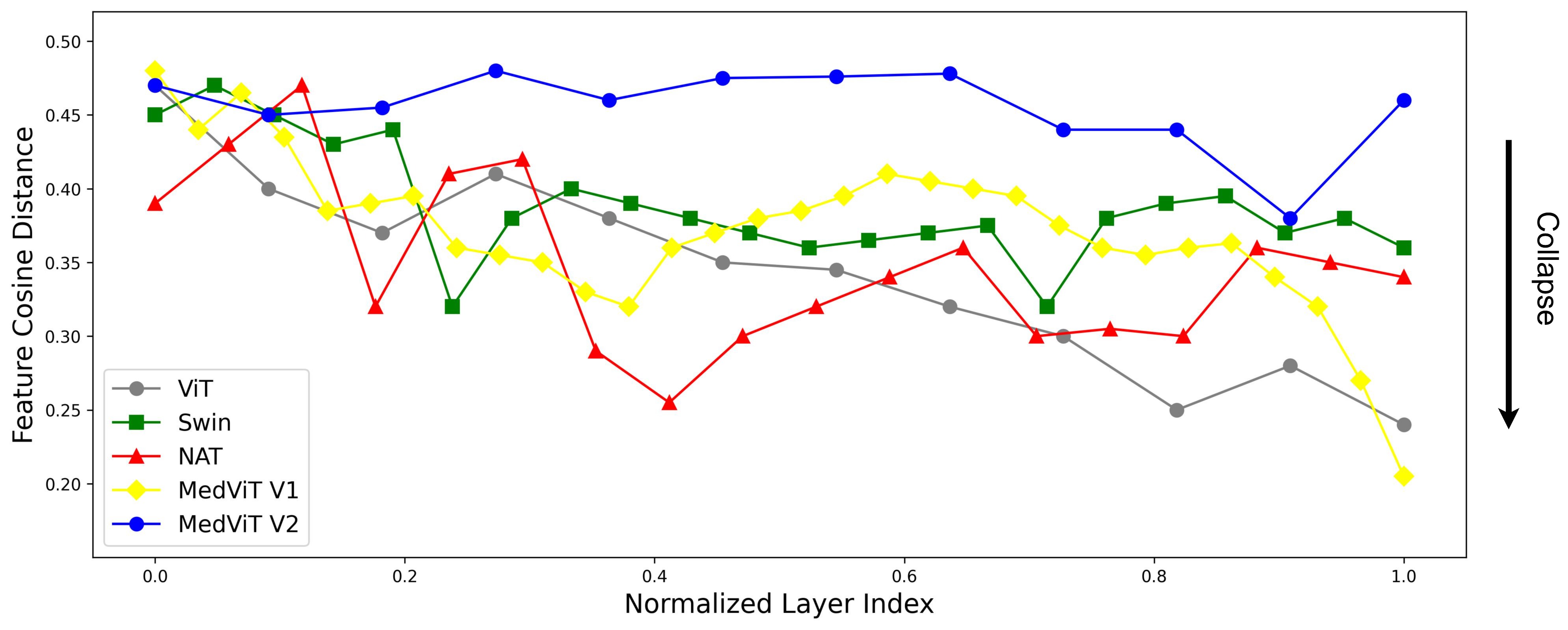}
\caption{
Comparative Analysis of Feature Cosine Distance. The plot illustrates the relationship between feature cosine distance values and normalized layer indices, accommodating models with varying total layer counts. A notable observation is the significant feature collapse behavior observed in the pre-trained MedViTV1 model. However, this phenomenon is confined to its deepest layers.}
\label{fig:feat_cos_dist}
\end{figure*}

\subsection{Feature Collapse Analysis}
Several recent works have mathematically analyzed the feature representation capabilities of deep models \cite{laurent2024feature, rangamani2023feature, woo2023convnext}. To quantitatively validate our observations, and building upon prior research, we conducted a feature cosine distance analysis. Given an activation tensor $X \in \mathbb{R}^{H \times W \times C}$, where $X_{i} \in \mathbb{R}^{H \times W}$ denotes the feature map of the $i$-th channel, we reshape each feature map into a $HW$-dimensional vector. We then compute the average pairwise cosine distance across all channels using the formula: $\frac{1}{ C^{2}} \sum_{i}^{ C} \sum_{j}^{ C} \frac{1-\cos( X_{i}, X_{j})}{2}$. In this context, a higher distance value signifies greater feature diversity, while a lower value indicates feature redundancy.

For this analysis, we randomly selected 5,000 images from various classes within the PathMNIST validation set. High-dimensional features were extracted from each layer of different models, including the pre-trained Swin model~\cite{liu2021swin}, the pre-trained ViT model~\cite{ViT}, the pre-trained NAT model~\cite{hassani2023neighborhood}, and the pre-trained MedViTV1~\cite{manzari2023medvit}. Subsequently, we computed the cosine distance per layer for each image and averaged these values across all images. The aggregated results are presented in Figure~\ref{fig:feat_cos_dist}. Consistent with our observations from earlier feature map visualizations, the pre-trained MedViTV1 model clearly exhibits a tendency towards feature collapse. This finding strongly motivates our decision to employ the DiNA attention mechanism to promote feature diversity during the learning process and effectively prevent feature collapse.

\section{Conclusion}
In this paper, we introduce a new family of hybrid models called MedViTV2, which combines enhanced Transformer blocks with KAN, resulting in significant performance improvements when scaled across various medical benchmarks. Additionally, MedViTV2 strikes an excellent balance between clean and robust accuracy on corruption benchmarks. Our experiments demonstrate that MedViTV2 achieves SOTA performance on all evaluated medical benchmarks. Notably, MedViTV2 reduces computational complexity by approximately 44\% compared to its prior version. This substantial improvement makes it highly suitable for real-time deployment and federated learning applications, addressing privacy concerns. As future work, we plan to apply our architecture as a backbone for medical segmentation and 3D medical imaging. We hope our study will inspire future research in realistic medical deployments.

%%Harvard
% \bibliographystyle{model2-names.bst}
% \biboptions{authoryear}
\bibliographystyle{elsarticle-num.bst}
% \biboptions{number,nonatbib}
\bibliography{refs}

\end{document}